\documentclass[journal]{IEEEtran}

\usepackage{amssymb}
\usepackage{amsmath}
\usepackage{color}
\usepackage{url}
\usepackage{cite}
\usepackage{array}
\usepackage{graphicx}
\usepackage{subfigure}
\usepackage{booktabs}
\usepackage{multirow}
\usepackage{threeparttable}
\usepackage[american]{babel}
\usepackage[pagebackref=true,breaklinks=true,citecolor=blue,colorlinks,bookmarks=false]{hyperref}

\ifCLASSINFOpdf

\else

\fi

\begin{document}

\title{Towards Real-time Eyeblink Detection in The Wild: \\Dataset, Theory and Practices}

	\author{Guilei~Hu,
		Yang~Xiao,
        Zhiguo~Cao,
        Lubin~Meng,
        Zhiwen~Fang,
        Joey Tianyi Zhou, and
        \\~Junsong~Yuan,~\IEEEmembership{Senior Member,~IEEE}
		\thanks{Guilei Hu, Yang Xiao, Zhiguo Cao and Lubin Meng are with National Key Laboratory of Science and Technology on Multi-Spectral Information Processing, School of Artificial Intelligence and Automation, Huazhong University of Science and Technology, China. E-mail: {guilei$\_$hu, Yang$\_$Xiao, zgcao, lubinmeng}@hust.edu.cn.}
        \thanks{Zhiwen Fang is with Guangdong Provincial Key Laboratory of Medical Image Processing, School of Biomedical Engineering, Southern Medical University, P. R. China. He is also with School of Energy and Mechanical-electronic Engineering, Hunan University of Humanities, Science and Technology, P. R. China.  E-mails: {fzw310@sina.com.}}
        \thanks{Joey Tianyi~Zhou is with Institute of High Performance Computing, A*STAR, Singapore. E-mail: {zhouty@ihpc.a-star.edu.sg.}}
		\thanks{Junsong Yuan is with the Computer Science and Engineering Department of University at Buffalo, the State University of New York, USA. E-mail: jsyuan@buffalo.edu.}
		\thanks{Yang Xiao is the corresponding author of this paper.}
		\thanks{Manuscript received February 19, 2014; revised September 17, 2014.}}
	
	\markboth{Journal of \LaTeX\ Class Files,~Vol.~11, No.~4, December~2012}%
	{Shell \MakeLowercase{\textit{et al.}}: Bare Demo of IEEEtran.cls for Journals}

\maketitle	

\begin{abstract}
Effective and real-time eyeblink detection is of wide-range applications, such as deception detection, drive fatigue detection, face anti-spoofing. Despite previous efforts, most of existing focus on addressing the eyeblink detection problem under constrained indoor conditions with relative consistent subject and environment setup. Nevertheless, towards practical applications, eyeblink detection in the wild is highly preferred, and of greater challenges. In this paper, we shed the light to this research topic. A labelled eyeblink in the wild dataset (i.e., HUST-LEBW) of 673 eyeblink video samples (i.e., 381 positives, and 292 negatives) is first established. These samples are captured from the unconstrained movies, with the dramatic variation on face attribute, head pose, illumination condition, imaging configuration, etc. Then, we formulate eyeblink detection task as a binary spatial-temporal pattern recognition problem. After locating and tracking human eyes using SeetaFace engine and KCF (Kernelized Correlation Filters) tracker respectively, a modified LSTM model able to capture the multi-scale temporal information is proposed to verify eyeblink. A feature extraction approach that reveals the appearance and motion characteristics simultaneously is also proposed. The experiments on HUST-LEBW reveal the superiority and efficiency of our approach. The comparisons with the existing state-of-the-art methods validate the advantages of our manner for eyeblink detection in the wild.
\end{abstract}

\begin{IEEEkeywords}
Eyeblink detection, eyeblink in the wild, spatial-temporal pattern recognition, LSTM, appearance and motion
\end{IEEEkeywords}

\IEEEpeerreviewmaketitle

\section{Introduction}

\IEEEPARstart{E}{yeblink} detection is of essential research value for the application scenarios of deception detection~\cite{Perelman2014Detecting}, drive fatigue detection~\cite{Bergasa2006Real}, face anti-spoofing~\cite{Pan2007Eyeblink}, dry eye syndrome recovery~\cite{Rosenfield2011Computer}, etc. During the past decades, numerous efforts~\cite{Ji2014Real,Wu2008Development,Dong2008Driver,Tabrizi2009Open,krolak2012eye,Chau2005Real,Hong2008Drivers,Morris2002Blink} have already been paid to this. Nevertheless, most of them are proposed without considering the case of eyeblink in the wild. Meanwhile, the existing eyeblink detection datasets \cite{Pan2007Eyeblink,drutarovsky2014eye,talkingface,radlak2015silesian} are generally captured under the constrained indoor conditions with the relative consistent subject and environment setup. However, towards some practical application scenarios eyeblink detection in the wild is more preferred. For instance, during the phase of deception detection the eyeblink visual data may be surreptitiously collected using the hidden cameras, under the unconstrained indoor or outdoor conditions~\cite{Perelman2014Detecting}. In this case, the effective and real-time eyeblink detection approach in the wild is essentially required to ensure the performance.

\begin{figure}[t]
\centering
\includegraphics[width=8.8cm]{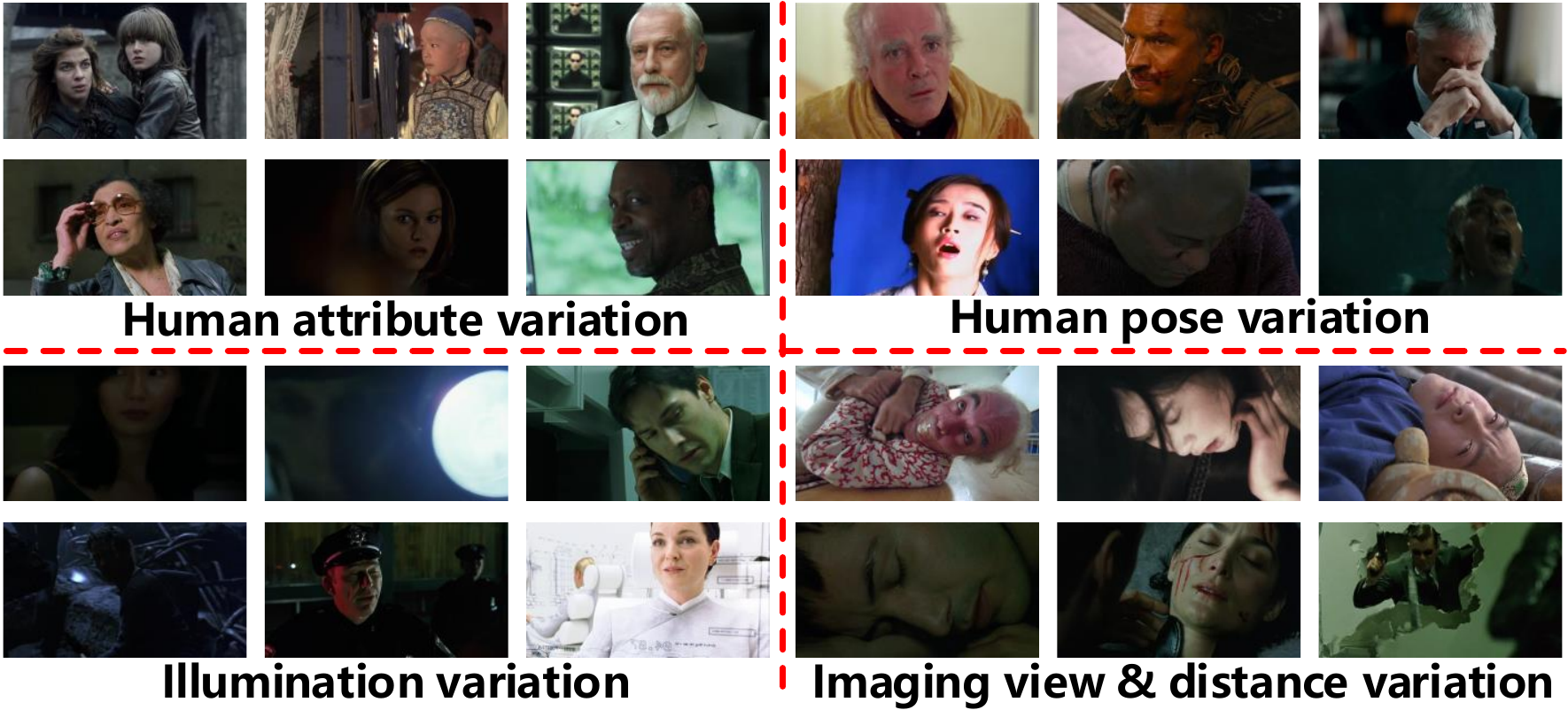}
\caption{The essential challenges towards eyeblink detection in the wild. The shown snapshots within HUST-LEBW dataset are captured from unconstrained movies.}
\label{fig:challenge}
\end{figure}

To this end, we first establish a challenging labelled eyeblink in the wild dataset termed HUST-LEBW. It consists of 673 eyeblink video clip samples (i.e., 381 positives, and 292 negatives) that captured from the unconstrained movies to reveal the characteristics of ``in the wild". Each positive sample covers one whole eyeblink process that corresponds to the eye status sequence of ``\texttt{eye open$\rightarrow$eye close$\rightarrow$eye open}". To our knowledge, HUST-LEBW \emph{is the first eyeblink in the wild dataset that involves the spatial-temporal sequence information.} Fig.~\ref{fig:challenge} shows some snapshots of the eyeblink samples within it. we can see that, there exits dramatic variations on human attribute, human pose, illumination, imaging viewpoint, and imaging distance. For instance, from the human attribute perspective the subjects involved in HUST-LEBW are of different ages, genders, races, skin colors and makeups. Meanwhile, the humans may or may not wear glass. This actually imposes great challenges to accurate eyeblink detection, both for eye localization and eyeblink verification.

\begin{table*}[t]
  \centering
  \scriptsize
  \caption{The attribute comparison among the proposed HUST-LEBW dataset and the existing eyeblink detection related datasets.}
  \begin{tabular}{@{\hspace{1.5mm}}c@{\hspace{1.5mm}}c@{\hspace{1.5mm}}c@{\hspace{1.5mm}}c@{\hspace{1.5mm}}c@{\hspace{1.5mm}}c@{\hspace{1.5mm}}c@{\hspace{1.5mm}}c@{\hspace{1.5mm}}c@{\hspace{1.5mm}}c@{\hspace{1.5mm}}
  c@{\hspace{1.5mm}}c@{\hspace{1.5mm}}}
    \toprule
    Dataset & Video clip amount &  Resolution & Person No. & Person race & Person age & Person sex & Person sight & Scene & Illumination & Imaging view & Imaging distance \\
    \midrule
        ZJU~\cite{Pan2007Eyeblink}   & 80  & 320$\times$240 & 20 & Asian & young & female & frontal & indoor & good  & frontal & fixed \\
      & (10877 frames)    &  &  & & middle-aged & male  & upward &  & stable &       &  \\
    \midrule
        Eyeblink8~\cite{drutarovsky2014eye}  & 8    & 640$\times$480 & 4 & Caucasian & young & female & frontal & indoor & good  & frontal &  fixed \\
     & (70992 frames)  &       &       &  & middle-aged & male  &       &       &       &       &  \\
    \midrule
       Talking face~\cite{talkingface} & 4   & 720$\times$576 & 1 & Caucasian & middle-aged & male  & frontal & indoor & good  & front & fixed \\
      & (5000 frames)  &       &       &       &       &       &   &    & stable &    &  \\
    \midrule
    Silesian5~\cite{radlak2015silesian} &  5  & 640$\times$480  & 5 & unknown  & unknown  & unknown & unknown & indoor  & good &  frontal & fixed\\
      &   (10877 frames)    &       &       &       &       &       &       &   &  stable  &  &\\
    \midrule
    Researchers' nights~\cite{researer}  & 107  & 640$\times$480 &  107 & Caucasian  & child &  female   & frontal & indoor  & variational & frontal & fixed \\
     & (224241 frames)      &    &  & & young & male &  &   &  &  &  \\
      &        &       &  &  & middle-aged &   &       &       &       &       &  \\
    \midrule
     HUST-LEBW  & 673  & 1280$\times$720 &  172 & Asian  & child &  female   & variational & indoor  & variational & variational & variational \\
     & (8749 frames)      &  1456$\times$600 &  &Caucasian & young & male &  & outdoor  &  &  &  \\
      &        &       &  & Melanoderm & middle-aged &   &       &       &       &       &  \\
          &             &  &  &  &elderly &       &       &       &       &       &  \\
    \bottomrule
    \end{tabular}%
  \label{table:dataset}%
\end{table*}

Next, \emph{we propose to formulate eyeblink detection in the wild task as a binary spatial-temporal pattern recognition problem}. In particular, a data-driven based real-time eyeblink detection approach that involves 2 stages of eye localization and eyeblink verification is proposed by us. During the spatial eye localization phase, the eye region is first detected using the off-the-shelf SeetaFace face parsing engine~\cite{Kan2017Funnel}, and then tracked by KCF (Kernelized Correlation Filters) tracker~\cite{Henriques2014High} to ensure the high running speed. Then towards eyeblink verification, Long Short Term Memory (LSTM) neural network is employed to model the temporal sequential procedure of eyeblink. Due to the issue that eyeblink may happen with the different time durations, we modify LSTM's architecture to consider the multi-scale temporal information of eyeblink.

Meanwhile, a feature extraction approach able to \emph{capture the appearance and motion information of eyeblink simultaneously} is also proposed. In particular, uniform Local Binary Pattern (LBP)~\cite{ahonen2006face} visual descriptor is extracted to reveal the appearance property of local eye region. And, the feature difference between the LBPs from 2 consecutive frames is used to encode the motion characteristics of eyeblink. The appearance and motion feature are concatenated as the input of LSTM.

Extensive experiments are then carried out on HUST-LEBW. The comparison with the state-of-the-art approaches demonstrates the superiority of our method on eyeblink detection in the wild , and its real-time running capacity.

The main contributions of this paper include:

$\bullet$ HUST-LEBW: the first eyeblink detection dataset that involves temporal sequential information towards ``in the wild" cases. It contains 673 video samples (i.e., 381 positives, and 292 negatives);

$\bullet$ A modified LSTM architecture abling to capture multi-scale temporal information of eyeblink is proposed;

$\bullet$ A uniform LBP-based eyeblink feature extraction method is proposed. It captures the appearance and motion information simultaneously.

HUST-LEBW and the source code of our work can be downloaded at \url{https://thorhu.github.io/Eyeblink-in-the-wild/}

The remaining of this paper is organized as follows. Sec.~\ref{sec:related_work} discusses the related work. The established HUST-LEBW dataset is introduced in Sec.~\ref{sec:HUST-LEBW}. Then, the proposed eyeblink detection method in the wild is illustrated in Sec.~\ref{sec:eyeblink_detection}. The essential implemetation details of the proposed eyeblink detection method are given in Sec.~\ref{sec:implementation}. Experiments and discussions are conducted in Sec.~\ref{sec:experiments}. Sec.~\ref{sec:conclusions} concludes the whole paper.

\begin{figure}[t]
\centering
\includegraphics[width=0.48\textwidth]{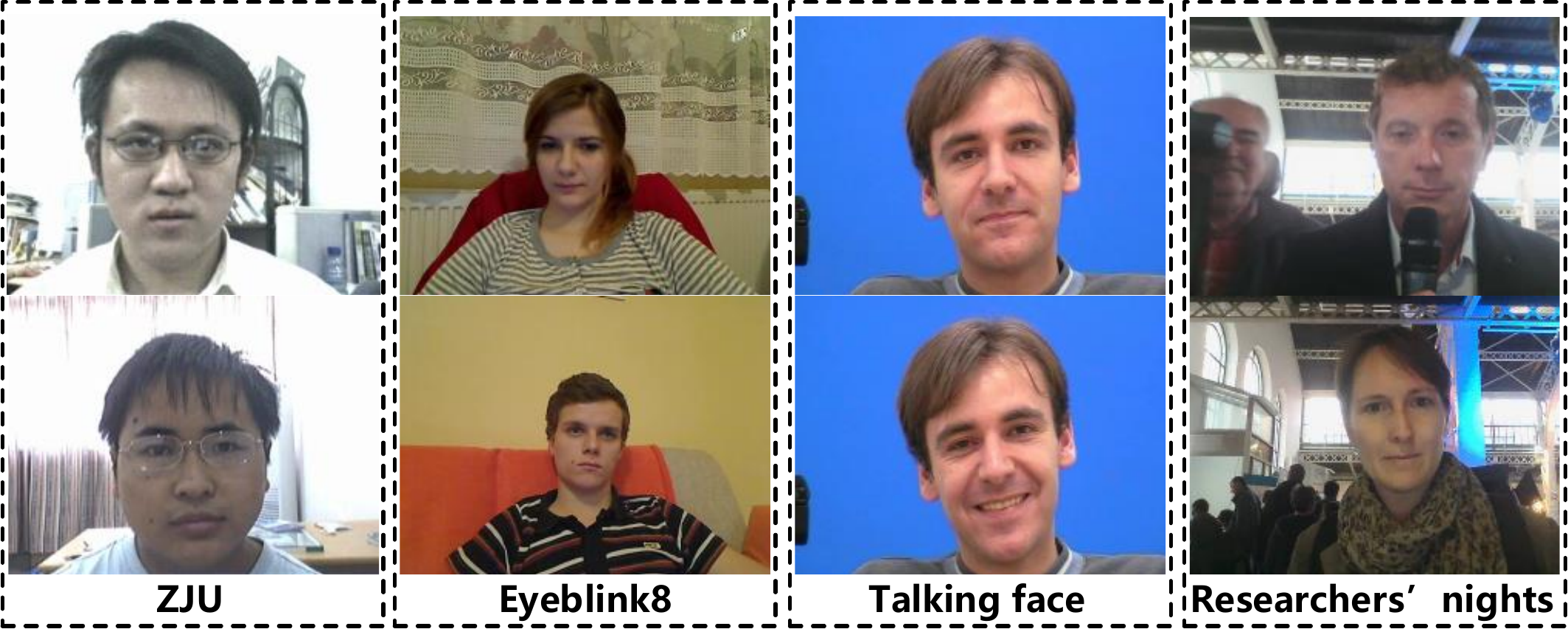}
\caption{Some eyeblink sample frames from the existing ZJU~\cite{Pan2007Eyeblink}, Eyeblink8~\cite{drutarovsky2014eye}, Talking face~\cite{talkingface}, and Researchers' nights datasets~\cite{researer}.}
\label{figure:otherdataset}
\end{figure}

\section{Related Work} \label{sec:related_work}

In this section, we will introduce and discuss the related work towards eyeblink detection in the wild in terms of \emph{dataset}, \emph{eyeblink verification} and \emph{eye localization} respectively.

\textbf{Eyeblink detection dataset.} Although numerous efforts have already been paid to address eyeblink detection problem, the available public datasets are still not abundant. ZJU~\cite{Pan2007Eyeblink}, Eyeblink8~\cite{drutarovsky2014eye}, Talking face~\cite{talkingface}, Silesian5~\cite{radlak2015silesian} and Researchers' nights~\cite{researer} are the representative ones with the spatial-temporal video information. Nevertheless, all of the 5 datasets above generally targets on the constrained indoor cases as shown in Fig.~\ref{figure:otherdataset}. The involved samples are captured from the limited number of volunteers, with the relatively consistent scene, subject, illumination and imaging setup. As consequence, they cannot reveal the ``in the wild" characteristics faced by some challenging application scenarios. And, the reported performance on these datasets is somewhat saturated (e.g., the detection rate of $99\%$ on ZJU and Silesian5). To facilitate the research on eyeblink detection in the wild, a more challenging dataset is indeed required. Accordingly, we propose to construct HUST-LEBW dataset in the way of collecting samples from the unconstrained live movies to essentially involve richer ``in the wild" eyeblink information. Compared to ZJU, Eyeblink8, Talking face Silesian5 and Researchers' nights~\cite{researer}, the samples in HUST-LEBW are of much higher diversity towards  scene, subject, illumination and imaging conditions. The detailed comparison among them is listed in Table~\ref{table:dataset} to verify this, in attributes of ``person number",  ``person race", ``person age", ``person sex", ``person sight", ``scene", ``illumination", ``imaging view", and ``imaging distance" respectively. Meanwhile, video clip amount and resolution is also listed. Hence, the severe attribute variation within HUST-LEBW will impose great challenges to accurate eyeblink detection.

\textbf{Eyeblink verification.} Towards the existing eyeblink verification approaches, we will introduce them from the perspectives of pattern recognition model, and feature extraction method respectively. First aiming to solve a binary pattern recognition problem, the existing eyeblink verification methods can be categorized into the~\emph{heuristic} and~\emph{data-driven} paradigms. Specifically, the heuristic way executes eyeblink verification mainly according to the pre-defined decision rules. For instance, when human face has been detected in advance a variance map of the sequential images is extracted to reveal the motion information in~\cite{Morris2002Blink}. Eyeblink verification is then carried out via executing thresholding operation on it, in spirit of computing the salient motion pixel ratio. Template matching is first executed to estimate the eye state in~\cite{krolak2012eye}. In the way of observing the correlation coefficient change in time, eyeblink is identified when the correlation coefficient is below a pre-defined threshold. KLT trackers are placed over the eye region to extract the motion information of eyeblink in~\cite{drutarovsky2014eye}. Eyeblink is consequently determined using the state machine with numerous of pre-defined threshold parameters. After acquiring the ``\texttt{open}" and ``\texttt{close}" status of eye using SVM, eyeblink is then confirmed according to the temporal contextual relationship between the resulting eye status in~\cite{Lee2010Blink}. With continuous eye tracking, eyeblink is recognized by observing whether the eyes are covered by eyelids in~\cite{torricelli2009adaptive}.  Actually, the effectiveness of most of these approaches above highly relies on the adaptability of the pre-defined thresholds for decision making. As consequence, they tend to be sensitive to subject and environment variation. To enhance the generalization capacity, some other researchers resort to data-driven manner. Being incorporated with the discriminative measures on eye status, Conditional Random Field (CRF) is employed to model the eyeblink procedure for verification in~\cite{Pan2007Eyeblink}. By extracting the EAR feature to characterize the eye opening degree using eye landmarks, SVM is finally used to verify the occurrence of eyeblink in~\cite{soukupova2016real}. Actually compared to the heuristic manner, data-driven approach is relatively seldom studied. And, our proposition falls into the data-driven paradigm to use LSTM framework with strong sequential information processing capacity to model the spatial-temporal procedure of eyeblink.

Besides the patter recognition model, another essential issue for eyeblink verification is feature extraction. Generally speaking, appearance feature (e.g., EAR~\cite{soukupova2016real}, LBP~\cite{Sun2009Robust}, Haar\cite{Liu2012}, or HOG~\cite{Dalal2005Histograms}) or motion feature (e.g., KTL tracker motion~\cite{soukupova2016real} or pixel-wise frame difference between the consecutive 2 frames~\cite{krolak2012eye}) are extracted to this end. Nevertheless, few approaches take appearance and motion information into consideration simultaneously. To address this, we propose to use uniform LBP as appearance feature and its difference between the consecutive 2 frames as motion feature to jointly characterize eyeblink.

\textbf{Eye localization.} Accurate eye localization is the key step for eyeblink detection within spatial domain. Some existing approaches~\cite{Wu2008Development,Tabrizi2009Open,Ji2014Real} resort to using color or spectral characteristics to locate eye. Another way is to use motion information~\cite{Tan2006Detecting} to detect and track eye. Nevertheless, their performance is not promising. Most of the state-of-the-art methods~\cite{krolak2012eye,Bradski2000The,soukupova2016real,DBLP:journals/corr/YinL17,zhang2016joint} resort to detect facial landmark to this end, in the way of face parsing. To achieve the balance between effectiveness and efficiency, we choose use SeetaFace engine~\cite{Kan2017Funnel} for eye detection first, and then track eye using KCF~\cite{Henriques2014High} for high running efficiency.

\section{HUST-LEBW : A Labelled Dataset for Eyeblink Detection in The Wild } \label{sec:HUST-LEBW}

\begin{figure}[t]
\centering
\includegraphics[width=0.48\textwidth]{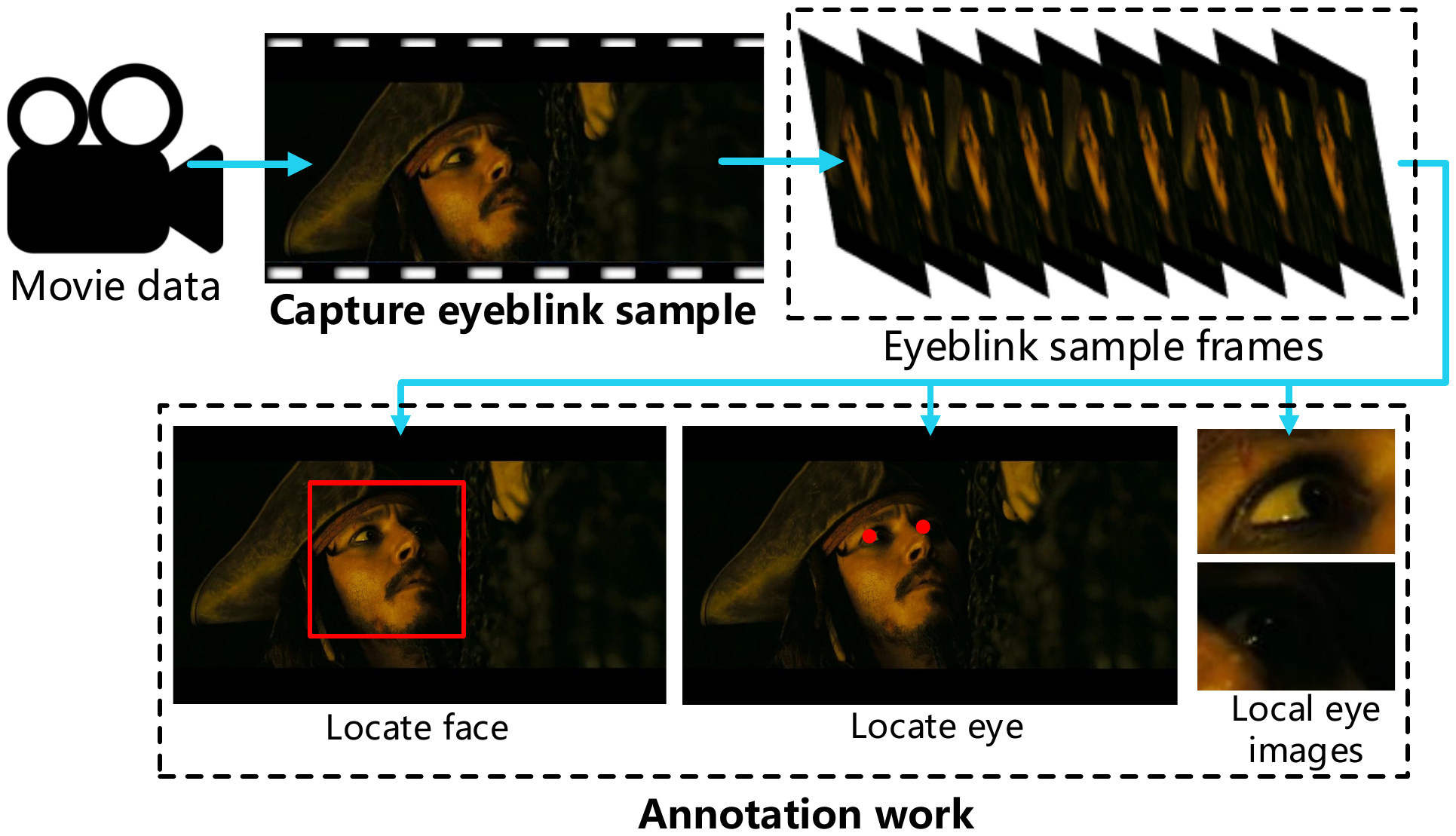}
\caption{The main construction pipeline of HUST-LEBW dataset.}
\label{figure:datasetpipeline}
\end{figure}

As shown in Fig.~\ref{fig:challenge}, eyeblink detection in the wild suffers from the challenges of variation on human attribute, human pose, illumination, imaging view and distance, etc. Nevertheless, the existing eyeblink detection datasets (e.g., ZJU~\cite{Pan2007Eyeblink}, Talking face~\cite{talkingface}, Eyeblink8~\cite{drutarovsky2014eye}, Silesian5~\cite{radlak2015silesian}, and Researchers' nights~\cite{researer}) cannot reveal the ``in the wild" characteristics well as indicated in Table~\ref{table:dataset} and Fig.~\ref{figure:otherdataset}. To address this, we propose to build a new labelled dataset for eyeblink detection in the wild (termed HUST-LEBW) to shed the light into this research field not well studied before. The essential difference between HUST-LEBW and the existing eyeblink detection datasets is that, we choose to collect eyeblink video clips from the unconstrained movies instead of from the limited number of volunteers under the indoor scene conditions. After capturing the eyeblink video clips from the movies, towards each frame the face region, point-wise eye location, and local eye region will be annotated as shown in Fig.~\ref{figure:datasetpipeline}. Next, we will illustrate the construction procedure and characteristics of HUST-LEBW in details.

\begin{table}[t]
  \centering
  \scriptsize
  \caption{The main attribute information of the 20 different moves for HUST-LEBW construction.}
  \label{tab:filmintro}%
    \begin{tabular}{@{\hspace{0.7mm}}c@{\hspace{0.7mm}}c@{\hspace{0.7mm}}c@{\hspace{0.7mm}}c@{\hspace{0.7mm}}c@{\hspace{0.7mm}}}
    \toprule
   Idx & Name & Filming location & Style  & Premiere time \\
    \midrule
    1 & A clockwork orange &  UK \& USA & Crime  \& thriller & 1971-12-19 \\
    2 & The last emperor & CN & Drama & 1987-10-23 \\
    3 & Farewell my concubine & CN & Drama  \& love & 1993-01-01 \\
    4 & Chungking express & CN &  Art  & 1994-07-14 \\
    5 & Leon  & FR \& USA & Action & 1994-09-14 \\
    6 & Ashes of time & CN & Emotional ethics & 1994-09-17 \\
    7 & The matrix & USA & Science fiction & 1999-04-30 \\
    8 & Dragon buster & CN & Costume & 2002-12-01 \\
    9 & The matrix reloaded & USA & Science fiction & 2003-05-15 \\
    10 & Pirates of the Caribbean & USA & Adventure \& magic & 2003-07-09 \\
    11 & Kill Bill 1 & CN \& USA & Action & 2003-10-10 \\
    12 & The lord of the rings3 & USA \& NZ & Fantasy \& action & 2003-12-01 \\
    13 & Blood diamond & USA \& DE & Adventure & 2006-02-06 \\
    14 & Memories of matsuko & JP & Drama \& music  & 2006-05-29 \\
    15 & The bourne ultimatum & USA & Action \& suspense & 2007-08-03 \\
    16 & Game of thrones & USA & War \& fantasy & 2011-04-17 \\
    17 & A Chinese fairy tale & CN & Fantasy \& love & 2011-04-19 \\
    18 & Black mirror & UK & Science \& thriller & 2011-12-04 \\
    19 & Mad max 4 & USA & Action & 2015-05-15 \\
    20 & Contratiempo & ES & Crime \& suspense & 2017-01-06 \\
    \bottomrule
    \end{tabular}%
\end{table}%

\subsection{Movie data source}

\begin{figure}[t]
\centering
\includegraphics[width=0.48\textwidth]{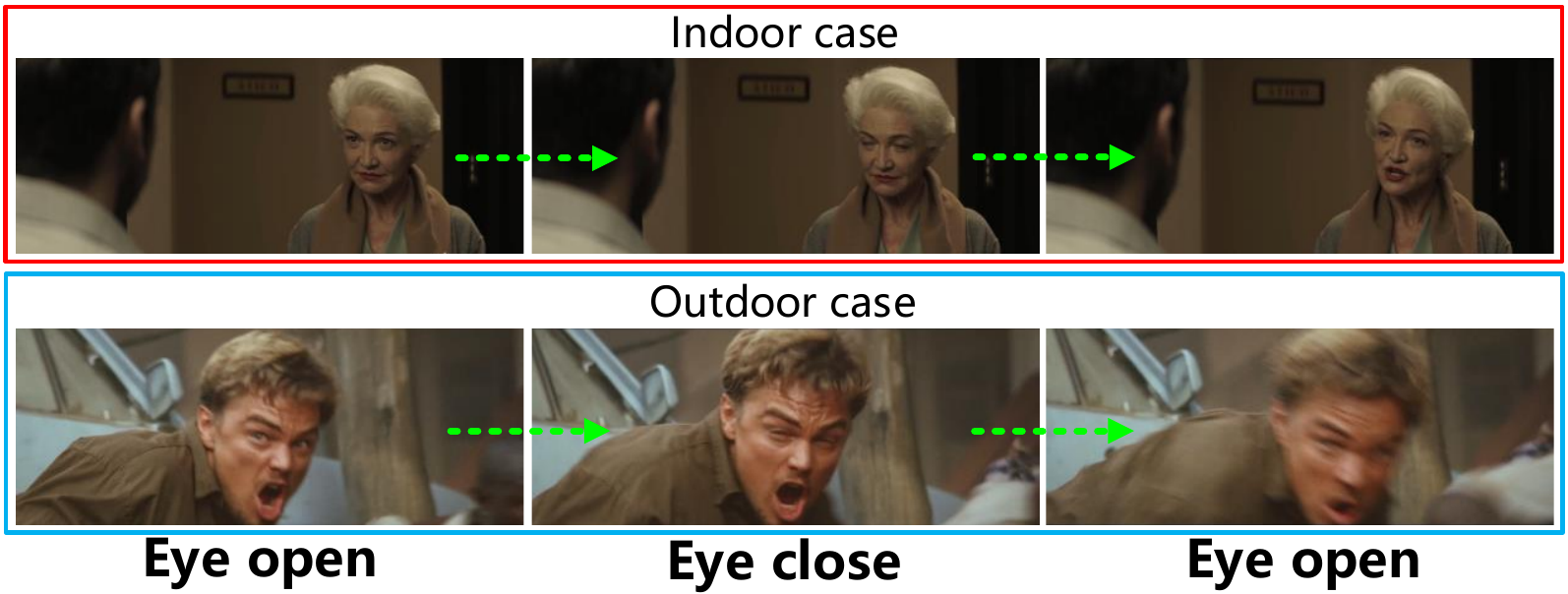}
\caption{The eyeblink video clip samples that correspond to the indoor and outdoor cases in HUST-LEBW dataset. Each eyeblink sample covers the whole eye status sequence of ``\texttt{eye open$\rightarrow$eye close$\rightarrow$eye open}". }
\label{figure:appendix b}
\end{figure}

\begin{figure}[t]
\centering
\includegraphics[width=0.48\textwidth]{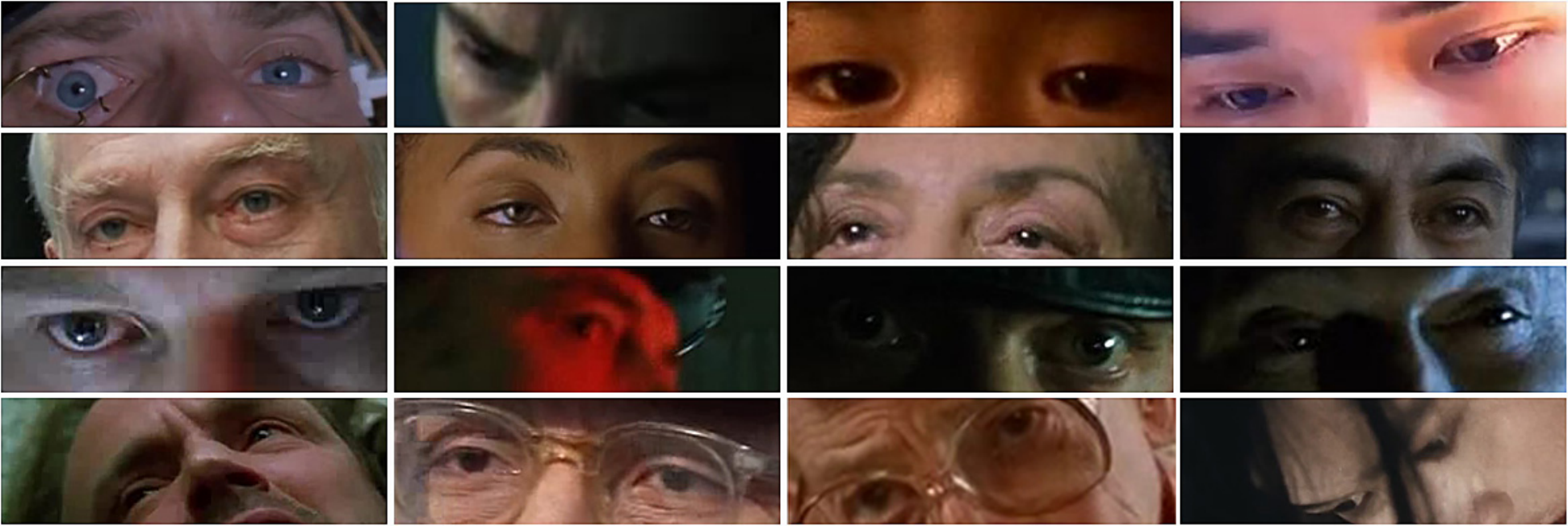}
\caption{The eye appearance variation among the 172 different persons within HUST-LEBW dataset.}
\label{figure:eyecolorshape}
\end{figure}

To reveal the ``in the wild" characteristics, the eyeblink samples in HUST-LEBW are collected from 20 different commercial movies. Their main attribute information (i.e., name, filming location, style and premiere time) is listed in Table~\ref{tab:filmintro}. It can be observed that, the attributes of these movies are actually of high diversity. Essentially, this helps to ensure the eventful ``in the wild" variation among the captured eyeblink samples in items of human attribute, human pose, scene / illumination condition, and imaging configuration as discussed in Table~\ref{table:dataset}. For instance, the employed 20 movies are shot in 8 countries from Asia, America, and Europe with the variational indoor and outdoor filming locations. Thus compared to the fixed indoor shooting condition of the existing eyeblink detection datasets~\cite{Pan2007Eyeblink,talkingface,drutarovsky2014eye,radlak2015silesian,researer}, acquiring eyeblink samples from these movies is of much stronger scene variation and challenges. Meanwhile, the discrepancy on movie style and premiere time also helps to promote the human attribute variation, which is more close to the practical applications. For example, the person races in HUST-LEBW include Asian, Caucasian and Melanoderm simultaneously. This actually cannot be met by the other datasets.

\subsection{Capture eyeblink in the wild sample}

\begin{figure}[t]
\centering
\includegraphics[width=0.48\textwidth]{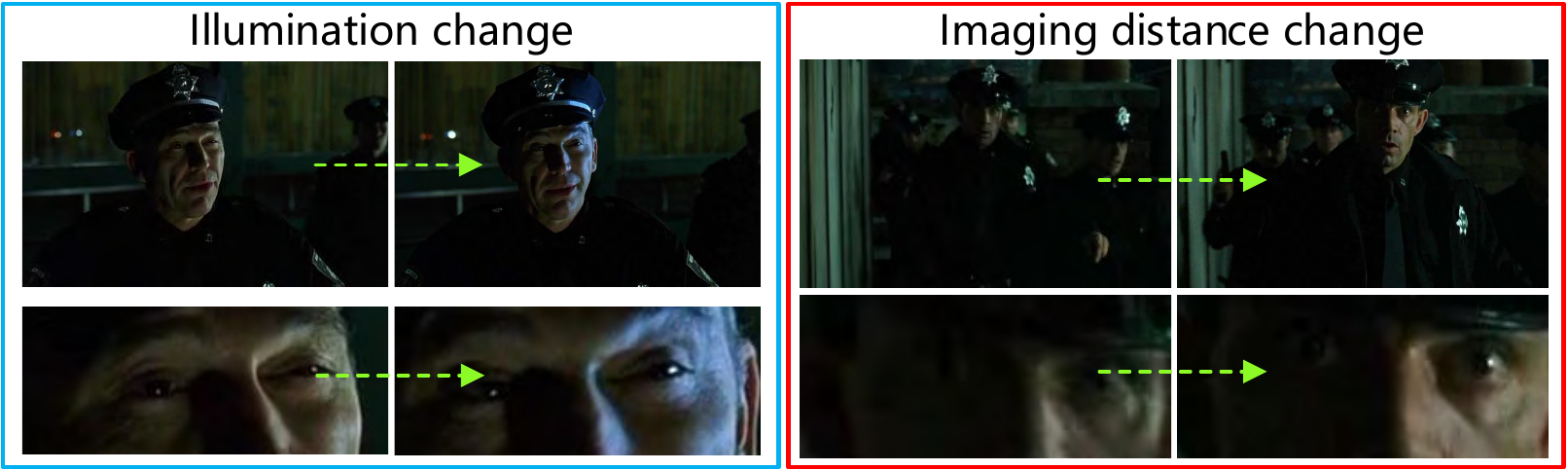}
\caption{The eye appearance variation that corresponds to the change on illumination and imaging distance within HUST-LEBW dataset. }
\label{figure:eyeillumidistance}
\end{figure}

From the 20 selected movies above, we then choose to capture the eyeblink in the wild samples in the form of video clip that covers the whole eye status sequence of ``\texttt{eye open$\rightarrow$eye close$\rightarrow$eye open}" as shown in Fig.~\ref{figure:appendix b}. Finally, we acquire 381 eyeblink video clips as the positive samples. Meanwhile, 292 non-eyeblink samples are also collected as the negative ones. As consequence, the yielded HUST-LEBW dataset consists of 673 samples in all (i.e., 381 positives, and 292 negatives).

Due to the high divergence of the employed movie data source, the captured eyeblink in the wild samples actually reveal dramatic variation on human attribute, human pose, scene condition, imaging view, and imaging distance as illustrated in Fig.~\ref{fig:challenge} and Table~\ref{tab:filmintro}.  These ``in the wild" factors essentially impose great challenges to effective eyeblink detection. For example, 172 persons of variational human attributes and poses are involved in HUST-LEBW dataset. Their eye appearance is actually of striking discrepancy as shown in Fig.~\ref{figure:eyecolorshape}. Meanwhile, even within the same eyeblink sample the eye appearance may also be of dramatic variation due to the change on illumination and imaging distance as shown in Fig.~\ref{figure:eyeillumidistance}. When concerning the variation of human attribute, human pose, scene and imaging condition simultaneously, accurately locating human eyes and characterizing the eye status for eyeblink detection in the wild is indeed not an easy task.

\begin{figure}[h]
\centering
\includegraphics[width=0.48\textwidth]{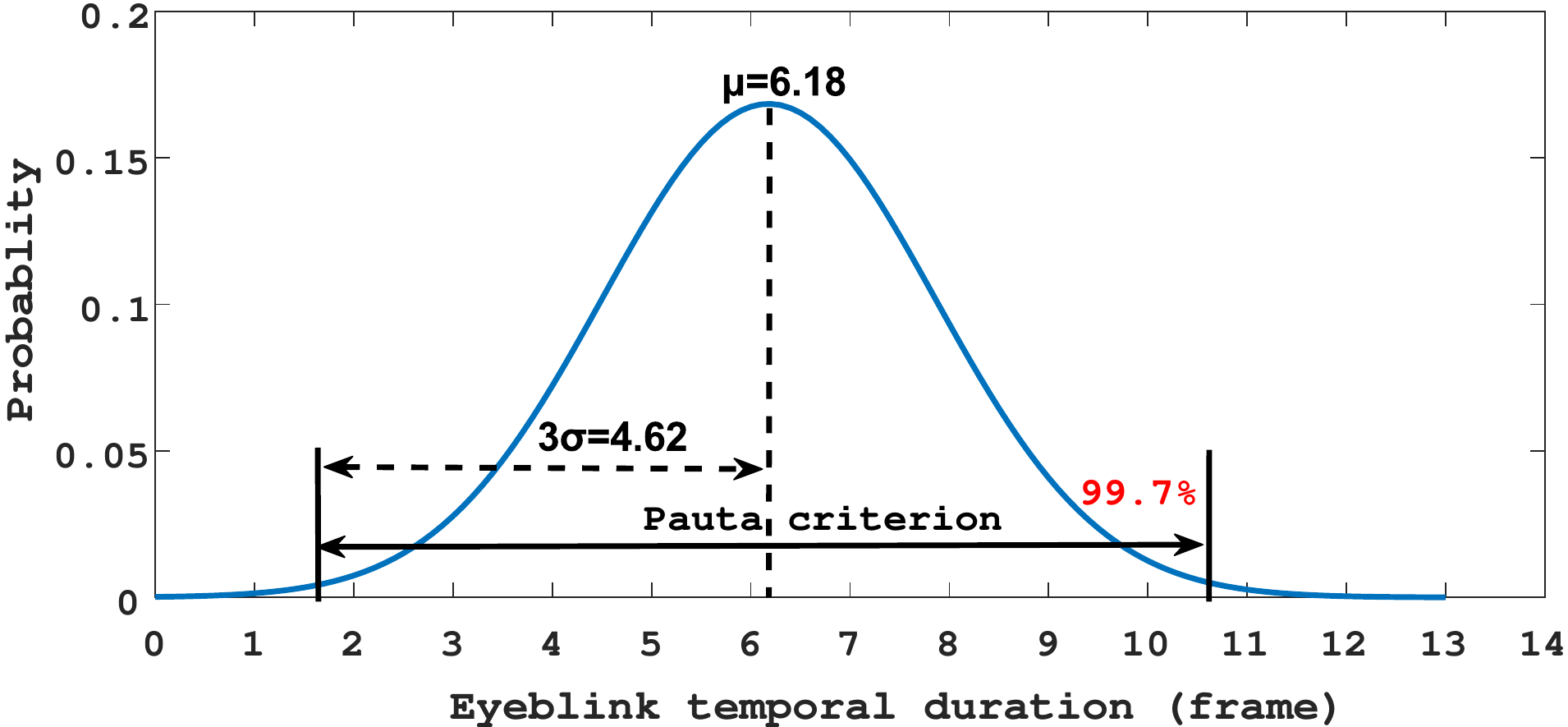}
\caption{The statistical result of temporal duration that corresponds to the 381 raw captured eyeblink video clips within HUST-LEBW dataset. The average frame rate is around 24 FPS.}
\label{figure:Pauta criterion}
\end{figure}

Since some existing eyeblink detection approaches (e.g.,~\cite{soukupova2016real}) and our proposed LSTM-based manner require the input eyeblink video clips to be of the same length, we choose to polish the raw captured eyeblink samples to be of the fixed temporal size. To this end, statistics on temporal duration of the raw eyeblink samples is executed as shown in Fig.~\ref{figure:Pauta criterion}. It can be observed that, the eyeblink temporal duration (frame) generally follows the Gaussian distribution with the mean value ($\mu$) of 6.18 and standard deviation ($\sigma$) of 1.54. To alleviate the outlier effect caused by human labelling bias, we set the fixed temporal duration of eyeblink sample as 10 frames according to the Pauta criterion (i.e., 3$\sigma$ criterion)~\cite{cao2017early} also as revealed in Fig.~\ref{figure:Pauta criterion}. In particular, during the eyeblink sample polish phase we will place the fully-closed eye frame around the middle of the eyeblink sample. Then, if the raw eyeblink sample is less than 10 frames the first and last frame will be copied uniformly for extension iteratively. Oppositely, if the raw eyeblink sample is more than 10 frames the excess frames will be cut from the left and right hand uniformly. Meanwhile since some eyeblink detection approaches (e.g.,~\cite{soukupova2016real}) require the input sample to be of 13 frames, we will also extend or cut the raw eyeblink samples to 13 frames to make HUST-LEBW dataset to be adapted to them.

\begin{figure}[t]
\centering
\subfigure[Face localization] {\label{face_location} \includegraphics[width=0.47\textwidth]{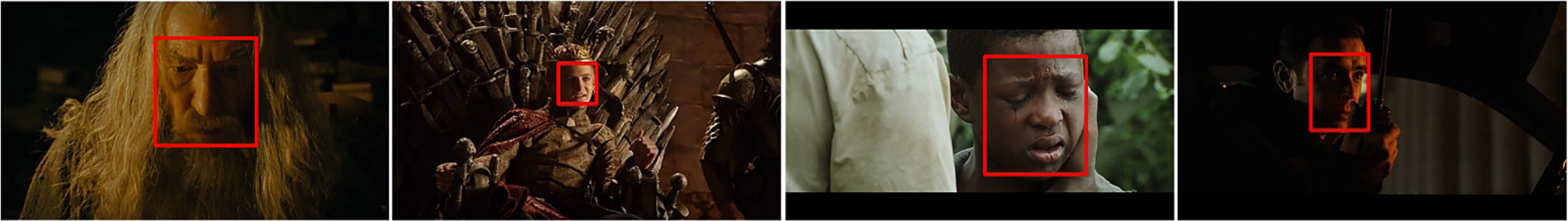}}  
\subfigure[Eye localization] {\label{eye_location} \includegraphics[width=0.47\textwidth]{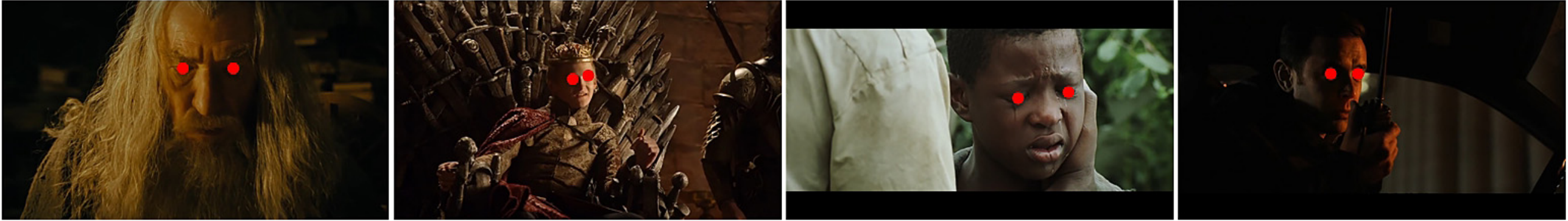}}
\subfigure[Local left and right eye image extraction] {\label{local_eye} \includegraphics[width=0.47\textwidth]{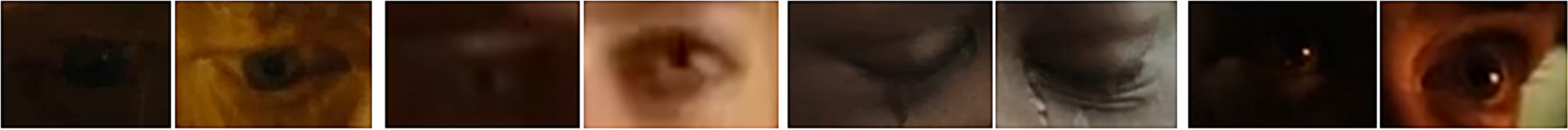}}
\caption{The examples of eyeblink sample annotation work on face localization, eye localization, and local eye image extraction.}
\label{fig:eyeblink_annotation}
\end{figure}

\subsection{Eyeblink sample annotation work} \label{sec:sample_annotation}

After acquiring the 673 eyeblink and non-eyeblink samples, we then execute annotation work on localizing face, localizing eye and extracting local eye images on each frame for performance evaluation towards practices. Next, we will introduce the annotation work in details.

\textbf{Face localization.} For each of the 8749 sample frames, we first use SeetaFace face parsing engine~\cite{Kan2017Funnel} to localize human face in terms of bounding box. Then, manual refinement is executed to ensure that the face bounding box can cover both of the right and left eye when they appear.

\textbf{Eye localization.} After face localization, we then manually localize the eye center at the point level frame by frame. If only one eye is visible, the coordinate of the invisible eye will be labelled as $\left(-1,-1\right)$.

\textbf{Local eye image extraction.} Using the acquired face bounding box and eye center position information, the local eye images are consequently extracted as follows. For one person, if both of the left and right eye are visible with labelled centers the height and width of the local eye image are calculated as
\begin{equation}
Eye_{hgt}=0.4 \times MH\left(P_{left},P_{right}\right),
\label{eq:eye_height}
\end{equation}
and
\begin{equation}
Eye_{wd}=0.4 \times MH\left(P_{left},P_{right}\right),
\label{eq:eye_width}
\end{equation}
where $P_{left}$ and $P_{right}$ indicate the position of left and right eye center; $MH\left(P_{left},P_{right}\right)$ represents the computation of Manhattan distance~\cite{stigler1986history} between $P_{left}$ and $P_{right}$. Meanwhile, if only one eye is visible the height and width will be determined using the face size information, following the principle proposed in~\cite{oguz1996proportion}. That is, the height and width of the local eye image are set as the $1/9$ of the face width. Some examples of eyeblink sample annotation are shown in Fig.~\ref{fig:eyeblink_annotation}.

It is worthy noting that, to ensure that the eyeblink sample annotation result is applicable to all the methods in experiments we will only localize the eyes and extract the local eye images visible for 13 frames. As consequence, we finally acquire 667 right eye samples and 644 left eye samples.

\subsection{Dataset split} \label{sec:split}

After the HUST-LEBW dataset has been built, we then split it into the training and test set. In particular, the training set consists of 448 samples. Among them, 254 samples are positives with 253 labelled right eyes and 243 labelled left eyes; 190 samples are negatives with 190 labelled right eyes and 181 labelled left eyes.

The test set consists of 225 samples. Among them, 127 samples are positives with 126 labelled right eyes and 122 labelled left eyes; 98 samples are negatives with 98 labelled right eyes and 98 labelled left eyes.

It is worthy noting that, the samples from the same movie will not appear in the training and test set simultaneously.


\begin{figure}[t]
\centering
\includegraphics[width=0.47\textwidth]{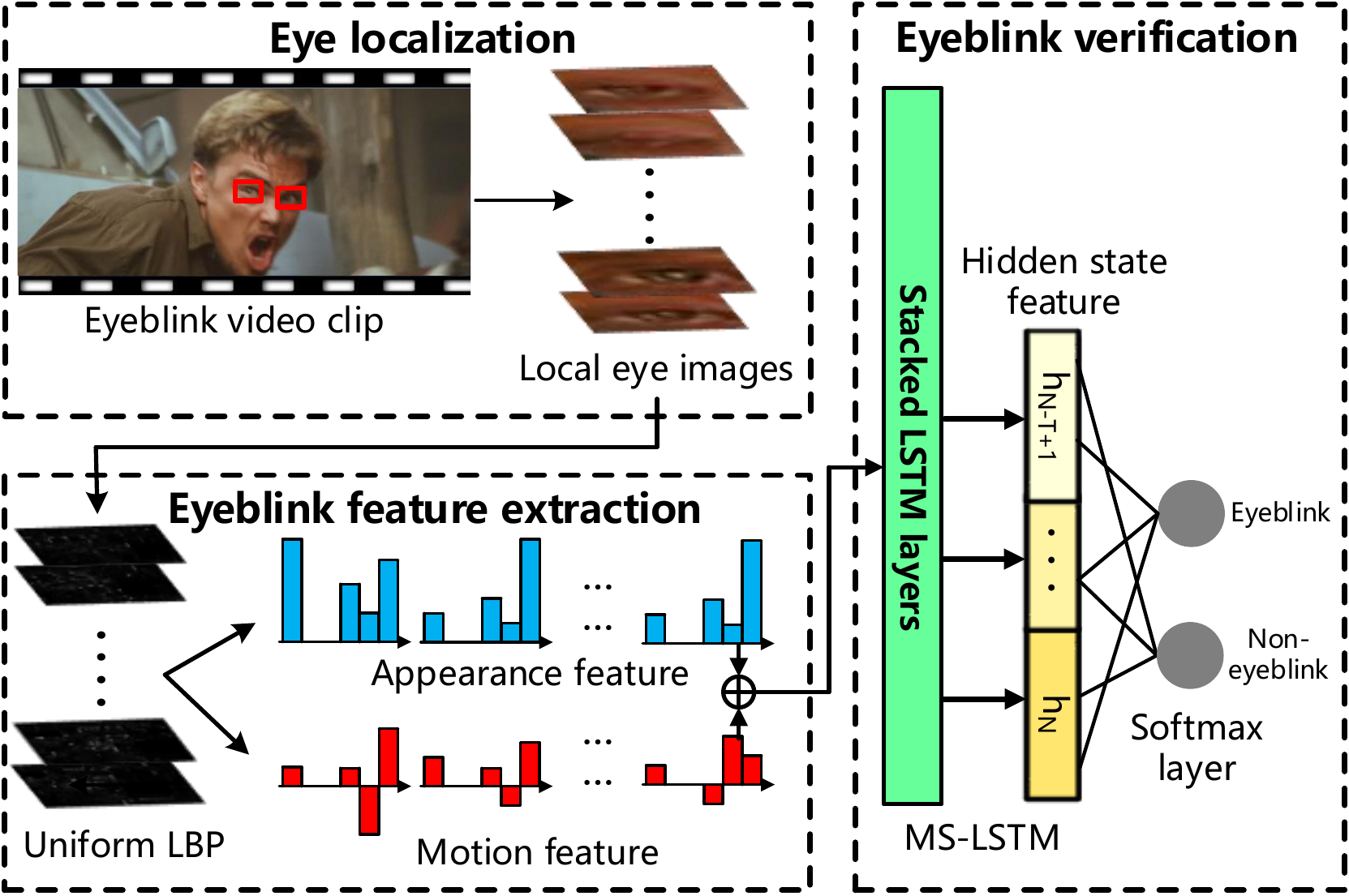}
\caption{The main technical pipeline of the proposed eyeblink in the wild detection approach.}
\label{figure:techpipeline}
\end{figure}

\section{Eyeblink in The Wild Detection Method : A Real-time Spatial-temporal Manner} \label{sec:eyeblink_detection}

To address eyeblink detection in the wild, eye localization is first executed at the spatial domain. Then, appearance and motion feature based on uniform LBP is simultaneously extracted per frame from the corresponding local eye images to characterize eyeblink. Multi-scale (MS) LSTM network able to handle multi-scale temporal information is consequently proposed to deal with the time series eyeblink characterization feature to address eyeblink verification. The main technical pipeline of the proposed eyeblink in the wild detection method is shown in Fig.~\ref{figure:techpipeline}. Next, we will illustrate it in details.

\begin{figure}[t]
\centering
\includegraphics[width=0.38\textwidth]{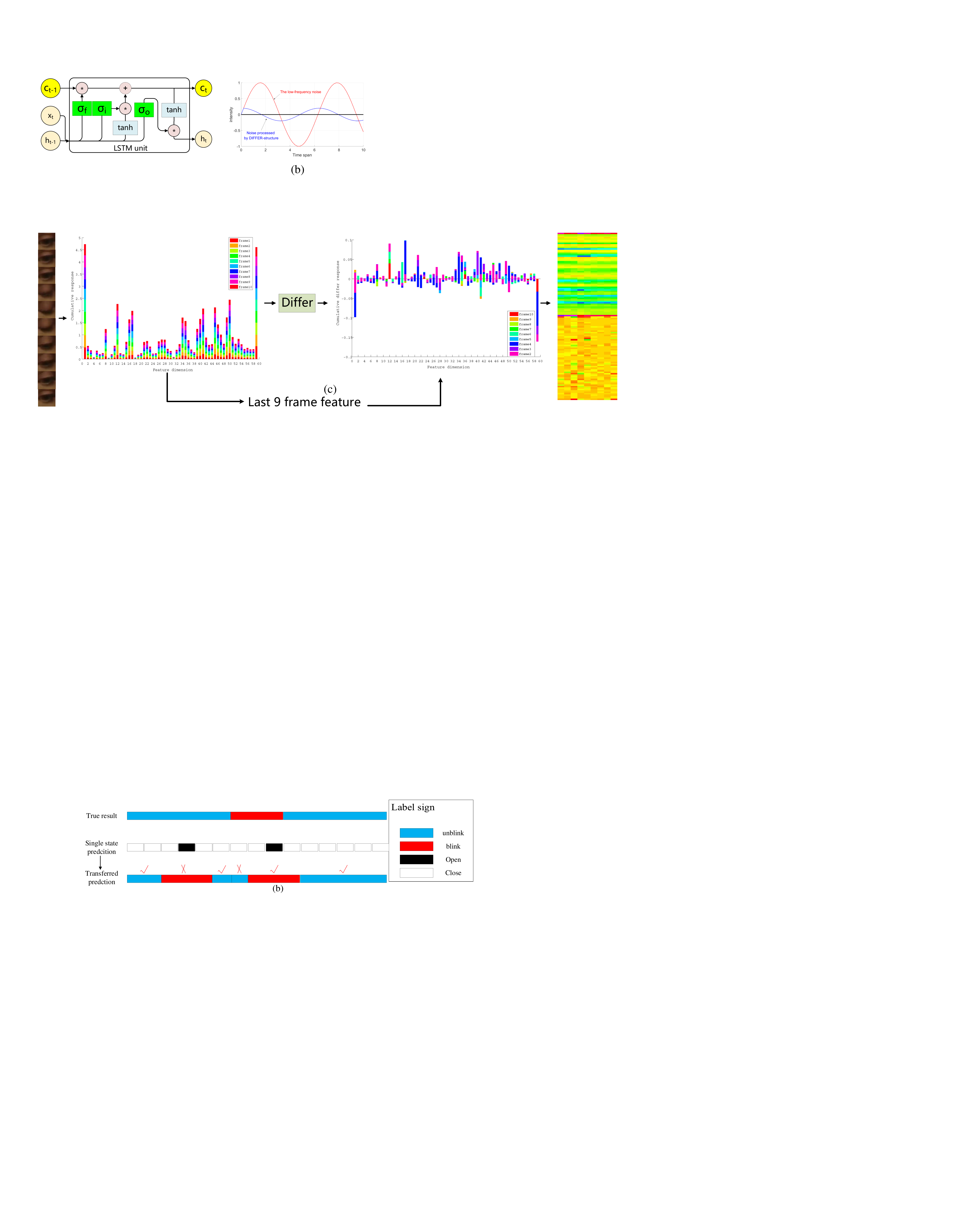}
\caption{The main structure of LSTM unit.}
\label{figure:lstmcell}
\end{figure}

\begin{figure}[t]
\centering
\includegraphics[width=0.48\textwidth]{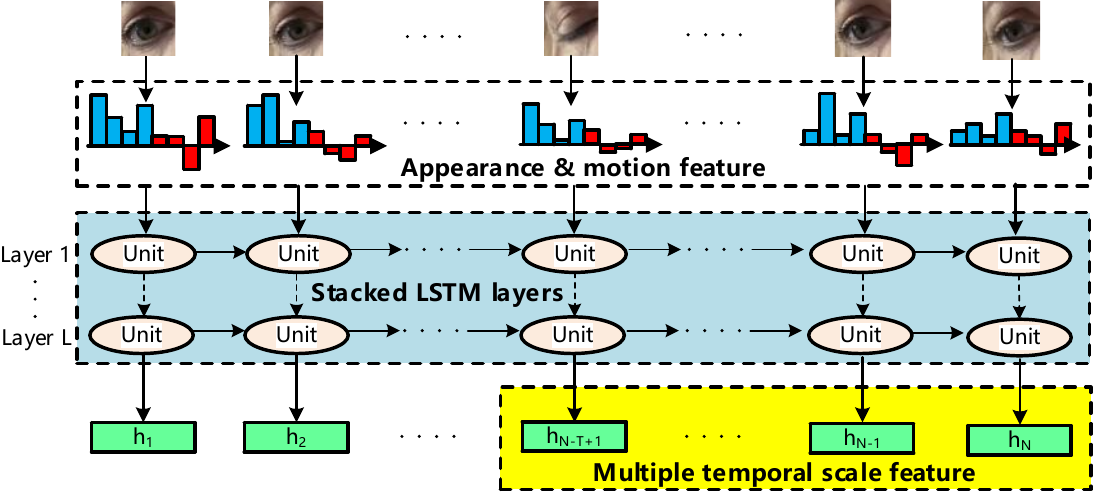}
\caption{The main structure of the proposed MS-LSTM model.}
\label{figure:mslstm}
\end{figure}

\begin{figure}[t]
\centering
\includegraphics[width=0.48\textwidth]{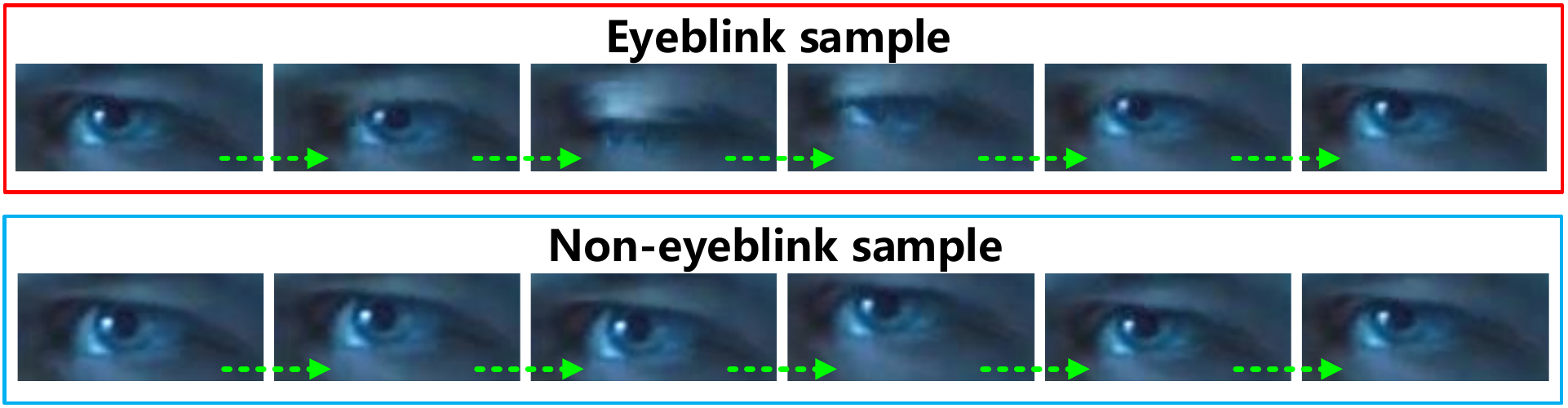}
\caption{The visual comparison between eyeblink from the same person.}
\label{figure:blink_unblink}
\end{figure}

\begin{figure}[t]
\centering
\subfigure[Left eye] {\label{left_distribution} \includegraphics[width=0.234\textwidth]{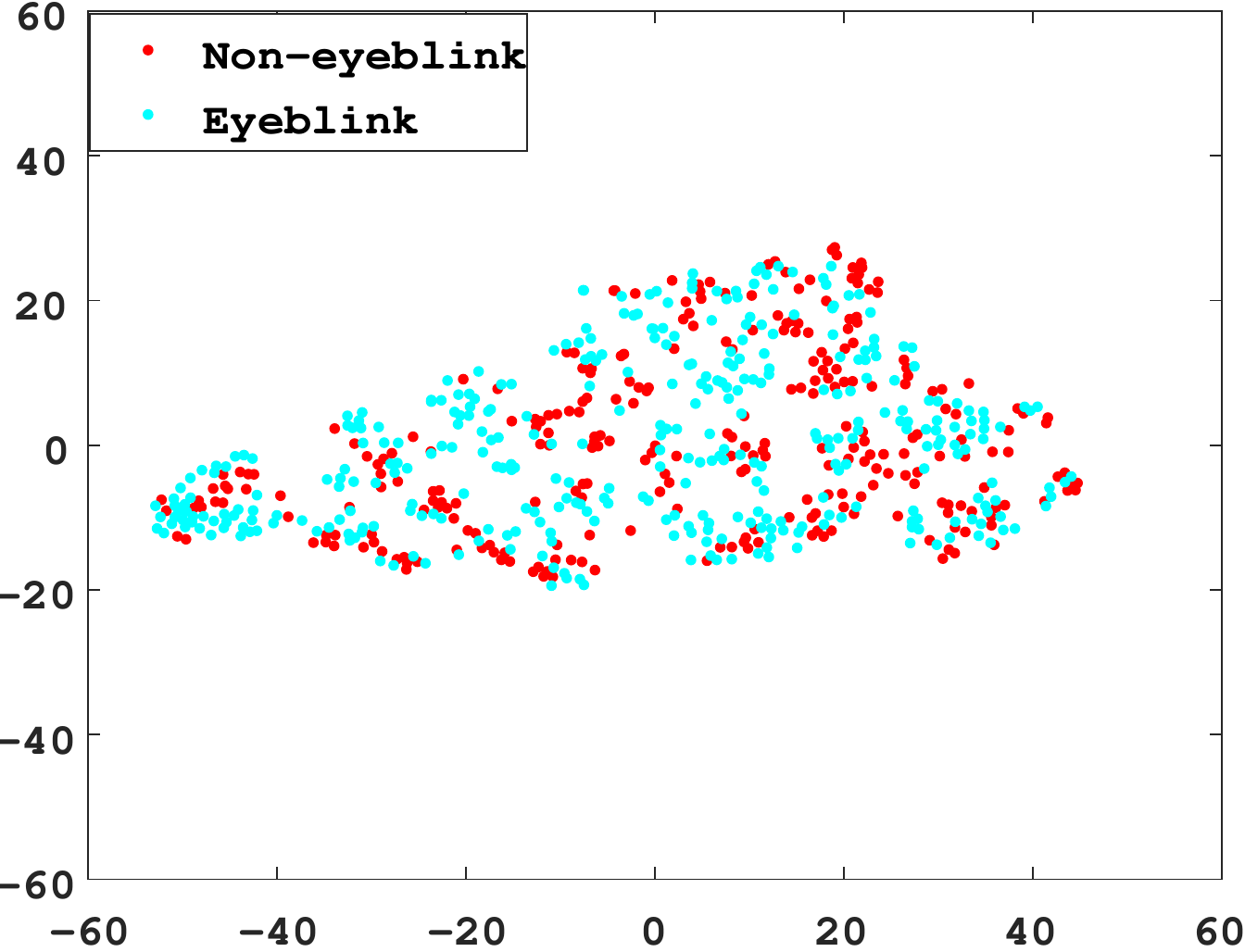}}  
\subfigure[Right eye] {\label{right_distribution} \includegraphics[width=0.234\textwidth]{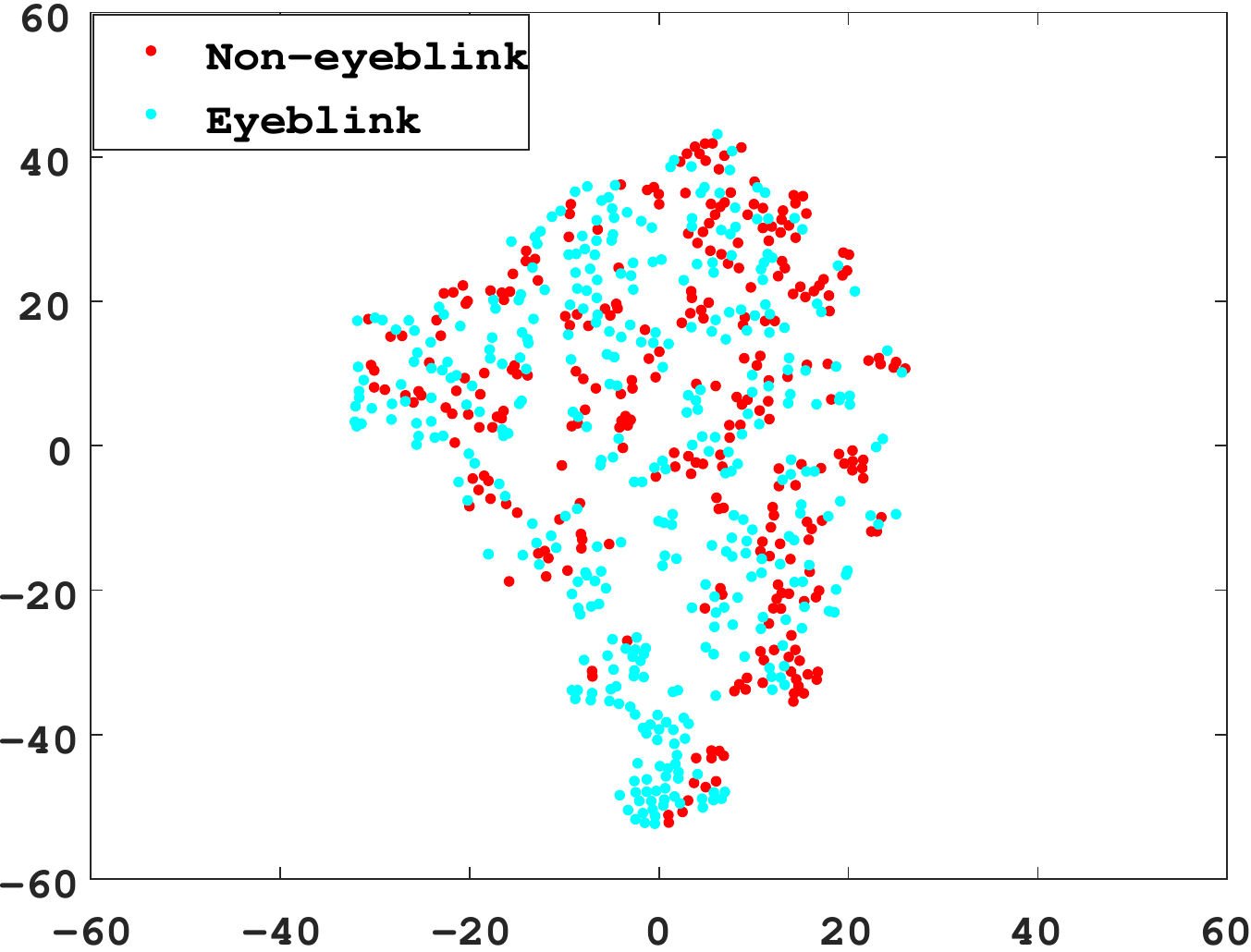}}
\caption{The feature distributions of the eyeblink and non-eyeblink samples within HUST-LEBW dataset, corresponding to the left and right eye respectively. They are drawn using t-SNE~\cite{maaten2008visualizing}. }
\label{fig:eyeblink_disreibution}
\end{figure}

\subsection{Eyeblink verification using multi-scale LSTM}

Eyeblink can be regarded as the facial activity that involves sequential eye statuses. Long Short-term Memory Network (LSTM)~\cite{hochreiter1997long} has been demonstrated to be one of the most successful deep learning models to deal with sequential data. It has already been applied to human body activity recognition~\cite{liu2017global} with promising performance. Inspired by this, we propose to apply LSTM to eyeblink verification.

LSTM is derived from Recurrent Neural Network (RNN)~\cite{hopfield1982neural} to model the long-term dependency within time series data. As shown in Fig.~\ref{figure:mslstm}, LSTM unit consists of a memory cell ($c_t$), an input gate ($\sigma_i$), a forget gate ($\sigma_f$), and an output gate ($\sigma_o$).  $\sigma_i$, $\sigma_f$, and $\sigma_f$ work collaboratively to prevent memory contents from being perturbed by irrelevant inputs and outputs to ensure long-term memory storage in $c_t$, in the way of controlling the information flow into and out of the LSTM unit. Meanwhile, the gradient vanishing and exploding problem met by RNN can also be alleviated in LSTM accordingly~\cite{hochreiter1997long}. However, intuitively applying the original LSTM model to eyeblink verification is not optimal. The insight is that eyeblink actually happens with the different temporal duration as revealed in Fig.~\ref{figure:Pauta criterion}, although they have been manually fixed to the same size within HUST-LEBW dataset. Essentially, the raw LSTM model cannot deal with the multiple temporal case within time series data well~\cite{hermans2013training}. To alleviate this, multi-scale LSTM (MS-LSTM) model is proposed by us from 2 perspectives as follows.

First instead of only using the output (i.e. the hidden state variable $h_t$ in Fig.~\ref{figure:mslstm}) of the last LSTM unit to be the input feature of softmax layer as for human body activity recognition~\cite{zhu2016co-occurrence,zhang2017geometric}, we choose to employ the outputs of the last $T$ LSTM units jointly by concatenation to involve richer multiple temporal scale information for eyeblink characterization.

Secondly inspired by the conclusion drawn in~\cite{hermans2013training} that the stacked RNN architecture can help to alleviate the multiple temporal scale problem, we transfer this idea to LSTM case by building $L$ stacked LSTM layers within MS-LSTM. Similar to stacked RNN~\cite{hermans2013training}, within the proposed MS-LSTM model the output of the previous LSTM layer will be employed as the input of the next LSTM layer in the parallel manner. Overall, the main structure of the proposed MS-LSTM model~\footnote{Within MS-LSTM, $L$ and $T$ are set as 2 using 3-fold cross-validation on training set.} is shown in Fig.~\ref{figure:mslstm}.


After the multiple temporal scale feature has been acquired within MS-LSTM, softmax layer will finally judge the type of input samples (eyeblink or non-eyeblink) as shown in Fig.~\ref{figure:techpipeline}. However, we argue that the original softmax loss~\cite{liu2017sphereface} is not discriminative enough for eyeblink verification since it is essentially a fine-grained visual recognition problem. To reveal this, we show one eyeblink sample and one non-eyeblink sample from the same person in Fig.~\ref{figure:blink_unblink}. It can be observed that, most of the frames within these 2 samples look similar except for the eye close part. This phenomenon may lead to the fact that, the eyeblink and non-eyeblink samples are not easy to distinguish in feature space. To further verify this, we exhibit the distribution of the eyeblink and non-eyeblink samples within HUST-LEBW dataset in Fig.~\ref{fig:eyeblink_disreibution}, using the appearance and motion feature illustrated in Sec.~\ref{sec:feature_extraction}. We can see that both in the left and right eye cases the eyeblink and non-eyeblink samples distribute with serious overlap, which is difficult to well discriminate. To enhance the discriminative power towards eyeblink verification, we propose to use the angular softmax (A-Softmax) loss~\cite{liu2017sphereface} with the promising performance for face verification. The intuition is that, face verification can also be regarded as a fine-grained visual recognition problem. Next, we will briefly introduce the key idea of A-Softmax loss.

\begin{figure}[t]
\centering
\subfigure[Original softmax loss] {\label{left_distribution} \includegraphics[height=3.5cm]{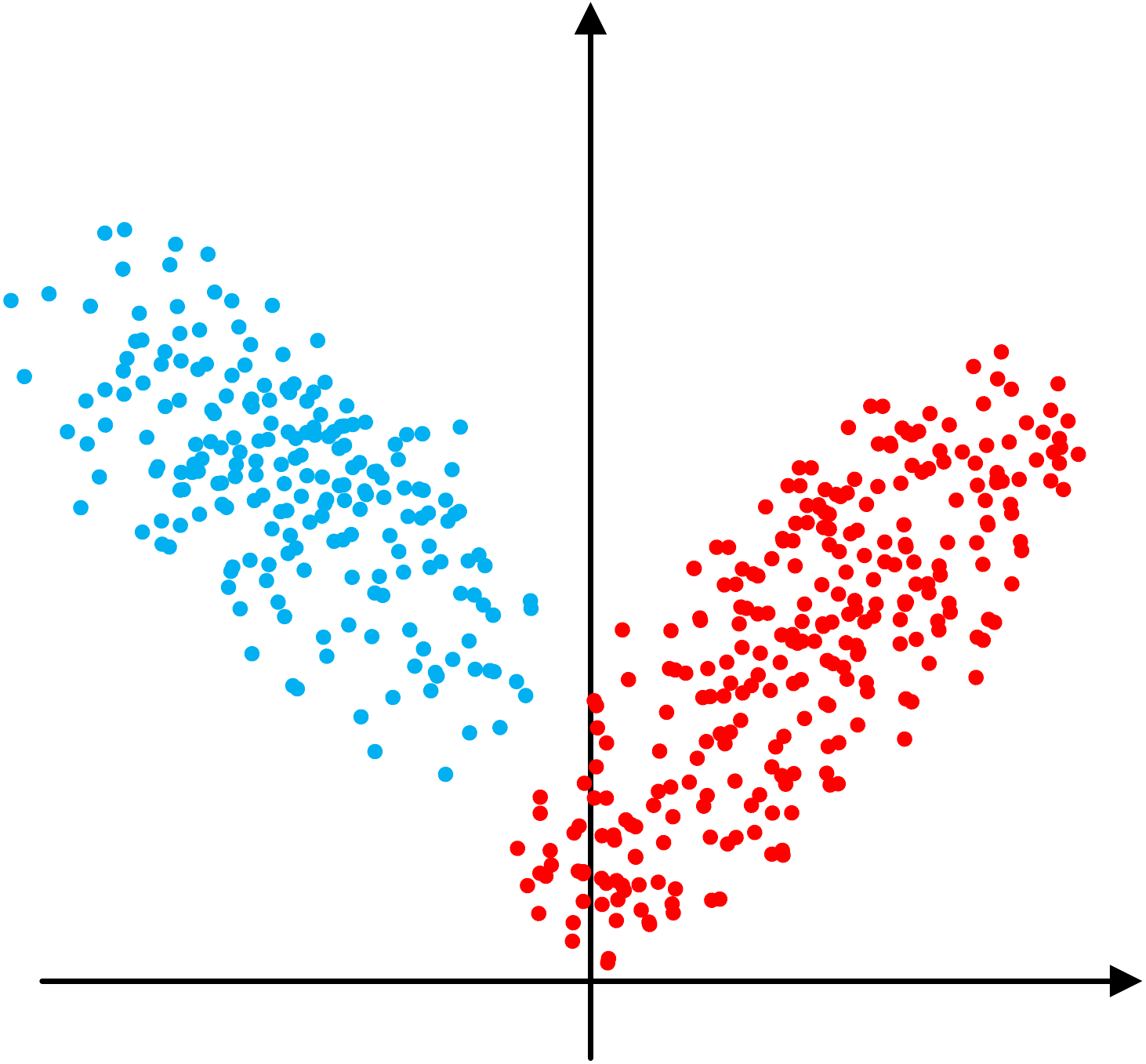}}  
\subfigure[A-softmax loss] {\label{right_distribution} \includegraphics[height=3.5cm]{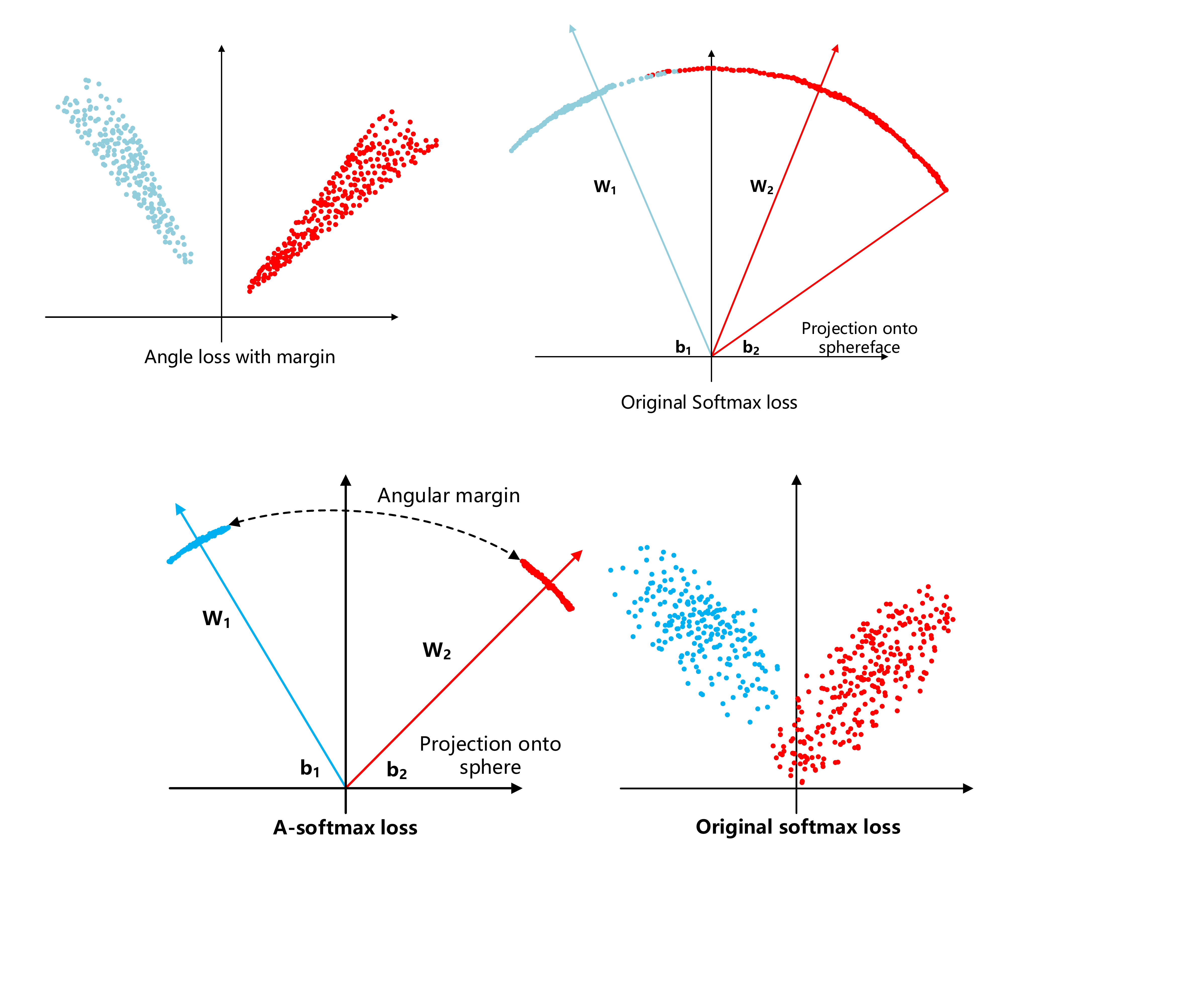}}
\caption{The visual comparison betweenthe  original softmax loss and A-softmax loss.}
\label{fig:softmax_a-softmax}
\end{figure}

For the binary pattern recognition problem of eyeblink verification, the decision boundary of the original softmax loss is defined as
\begin{equation}
\left(\textbf{W}_1-\textbf{W}_2\right)x+b_1-b_2=0,
\end{equation}
where $x$ indicates the input feature vector; $\textbf{W}_i$ and $b_i$ represent the weights and bias. With the constrain of $\left\|\textbf{W}_1\right\|=\left\|\textbf{W}_2\right\|=1$ and $b_1=b_2=0$, the decision boundary will be
\begin{equation}
\left\|x\right\|\left(cos\left(\theta_1\right)-cos\left(\theta_2\right)\right)=0,
\label{eq:modified_softmax}
\end{equation}
where $\theta_i$ is the angle between $\textbf{W}_i$ and $x$. As a result, the new 2-class decision boundary is only related to $\theta_i$. Actually, the modified softmax loss in Eqn.~\ref{eq:modified_softmax} enables the neural network to learn the angle-based decision boundary. However, it cannot ensure the strong discriminative power and generalization capacity. To alleviate this, A-Softmax loss introduces a integer $m$ $\left(m\geq1\right)$ to control angular margin between the 2 classes. Accordingly, the decision boundaries for the 2 classes are defined as
\begin{equation}
\left\|x\right\|\left(cos\left(m\theta_1\right)-cos\left(\theta_2\right)\right)=0,
\label{eq:a_softmax1}
\end{equation}
and
\begin{equation}
\left\|x\right\|\left(cos\left(\theta_1\right)-cos\left(m\theta_2\right)\right)=0,
\label{eq:a_softmax2}
\end{equation}
respectively. In summary, A-Softmax loss is to project the samples from Euclidean feature space to angular feature space and guarantees the angular margin between the 2 classes as shown in Fig.~\ref{fig:softmax_a-softmax}. In this way, the discriminative power and generalization capacity can be enhanced towards fine-grained eyeblink verification task. The detailed definition of A-Softmax loss can be found in~\cite{liu2017sphereface}.

\begin{figure}[t]
\centering
\subfigure[Appearance feature] {\label{appearance} \includegraphics[height=2.68cm]{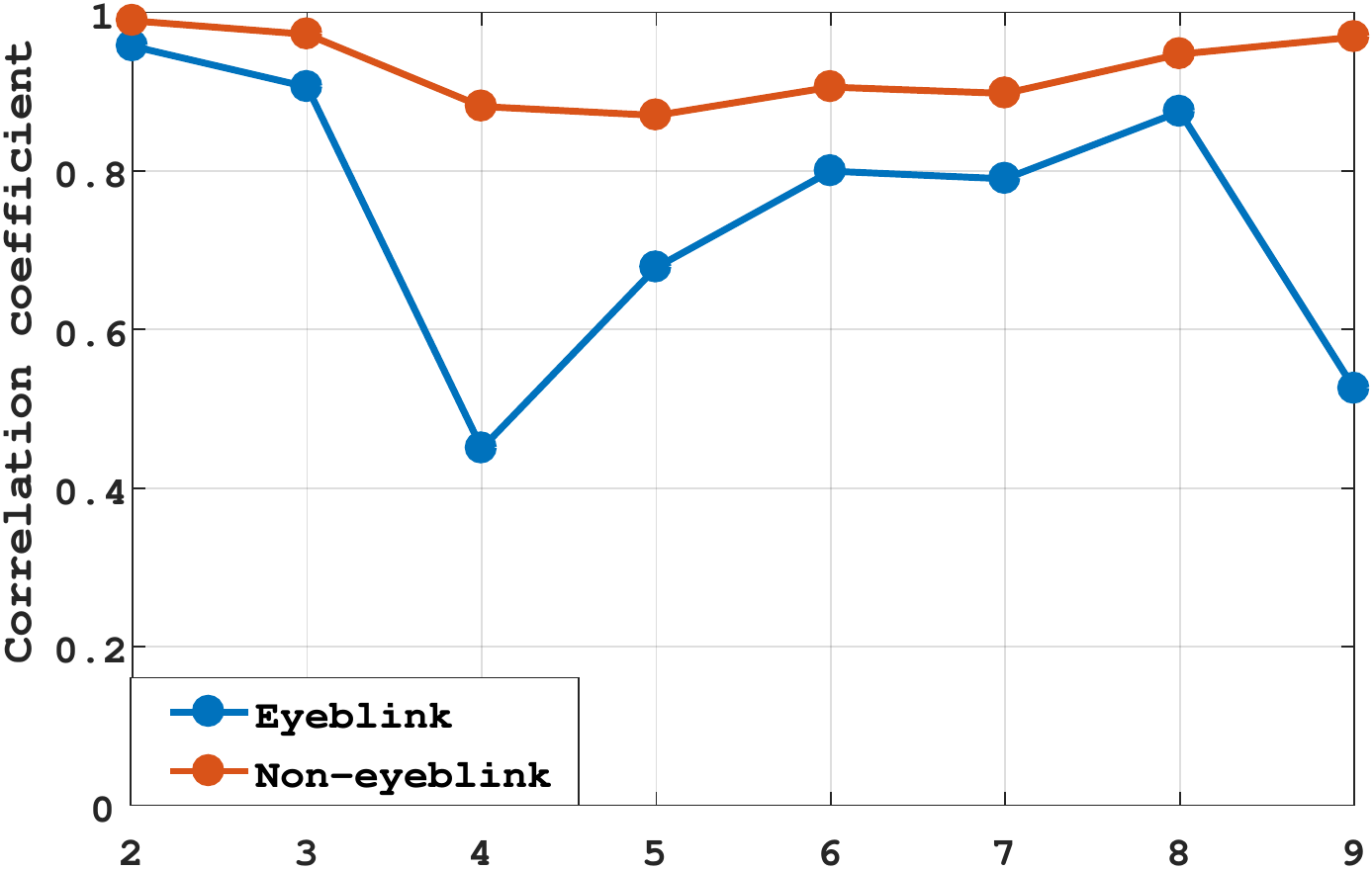}}  
\subfigure[Motion feature] {\label{motion} \includegraphics[height=2.68cm]{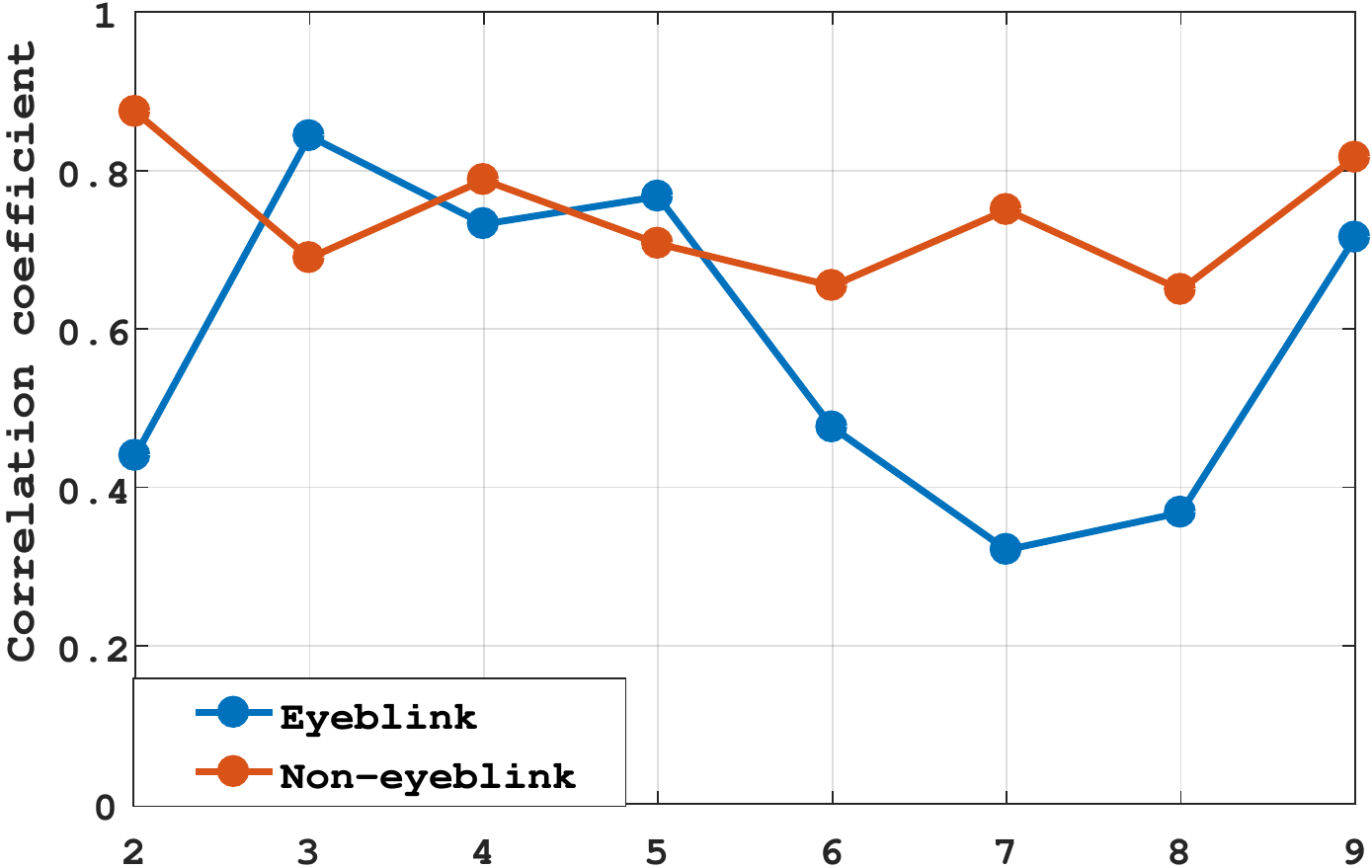}}
\caption{The correlation coefficients between the current and the next frame from the appearance and motion feature perspectives respectively, corresponding to the eyeblink and non-eyeblink samples shown in Fig.~\ref{figure:blink_unblink}.}
\label{figure:correlation}
\end{figure}


\subsection{Low-level appearance and motion feature extraction for eyeblink characterization}~\label{sec:feature_extraction}
Inspired by the two-stream (i.e., appearance and motion stream) human body activity recognition paradigm~\cite{simonyan2014two}, we propose to extract low-level appearance and motion feature simultaneously per frame as the input of MS-LSTM for eyeblink characterization. Concerning the real-time running issue, the lightweight uniform LBP visual descriptor~\cite{ahonen2006face} is used instead of the high-cost deep Convolutional Neural Network (CNN)~\cite{lecun2015deep} and optical flow~\cite{brox2011large} as in~\cite{simonyan2014two}. Another main reason for why we use uniform LBP is that it is rotation-insensitive~\cite{ojala2002multiresolution}, which is beneficial for eyeblink verification in the wild. As shown in Fig.~\ref{figure:eyecolorshape}, the eyeblink in the wild samples are often of different rotation angles due to the variational human poses or imaging views as revealed in Fig.~\ref{fig:challenge}.

Specifically, towards each frame uniform LBP  is extracted from the local eye image as the appearance feature. Besides, we also propose to calculate the difference between the uniform LBPs from 2 consecutive frames as the motion feature to reveal the eye status evolution during eyeblink. Intuitively, the appearance and motion feature is of the same dimensionality. They are concatenated as the input of MS-LSTM for spatial-temporal eyeblink characterization, corresponding to each frame except the first one.

To reveal the discriminative capacity of the proposed feature extraction method, the uncentered feature correlation coefficient~\cite{lee1988thirteen} between the current and the next frame towards the eyeblink and non-eyeblink samples in Fig.~\ref{figure:blink_unblink} is calculated as
\begin{equation}
corr\left(\mathbf{f_c,} \mathbf{f_n}\right)=\frac{\mathbf{f_c}\cdot\mathbf{f_n}}{\left\|\mathbf{f_c}\right\|\left\|\mathbf{f_n}\right\|},
\label{eq:corr_coeff}
\end{equation}
where $\mathbf{f_c}$ and $\mathbf{f_n}$ indicate the extracted eyeblink feature vector from the current and the next frame. The correlation coefficients of the different frames are shown in Fig.~\ref{figure:correlation}, from the appearance and motion feature perspectives respectively. We can see that, generally the frames from the eyeblink sample are of lower correlation coefficients. Meanwhile, the non-eyeblink sample frames possess the relatively consistent correlation coefficients. The phenomena above reveals that, our feature extraction approach can essentially capture the dynamic appearance and motion characteristics within eyeblink.

\subsection{Local eye image extraction} \label{sec:local-eye}
\begin{figure}[t]
\centering
\includegraphics[width=0.485\textwidth]{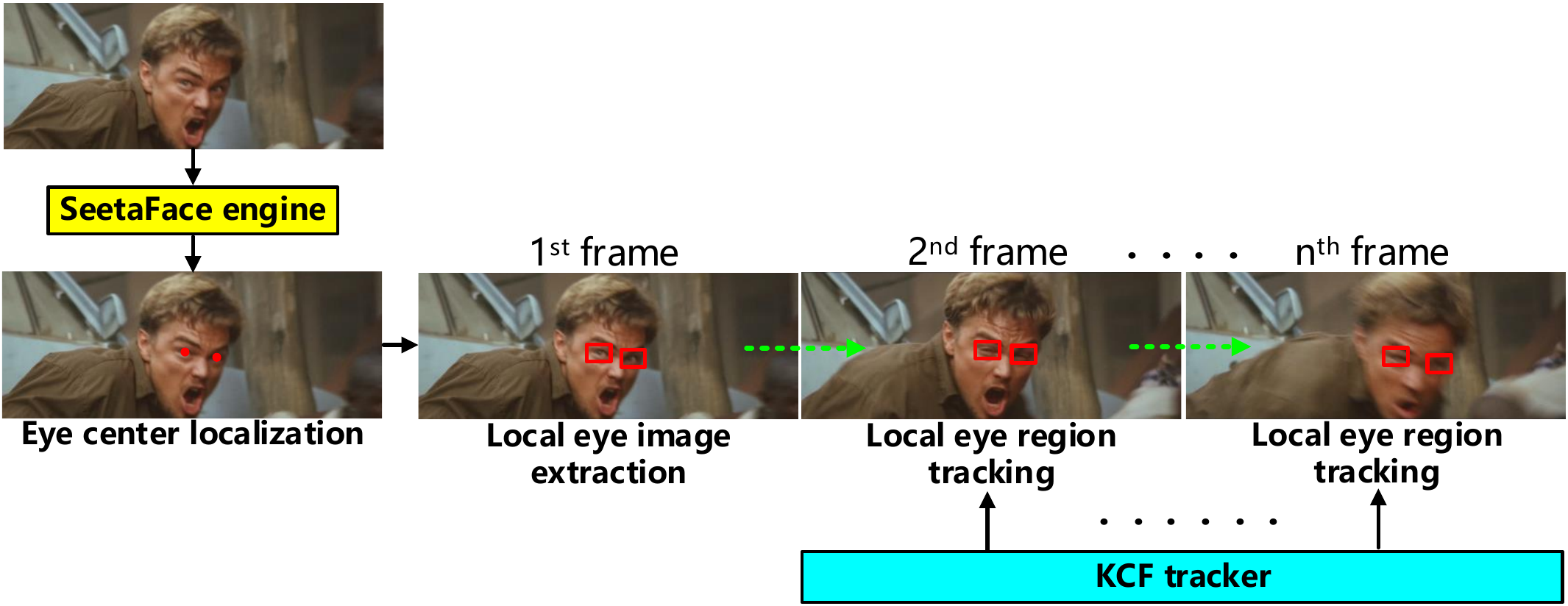}
\caption{The main technical pipeline for local eye image extraction.}
\label{figure:local_eye_extraction}
\end{figure}

As illustrated in Fig.~\ref{fig:eyeblink_annotation} and Sec.~\ref{sec:feature_extraction}, appearance and motion feature is extracted from the local eye images for eyeblink characterization. Thus, the effective and efficient local eye image extraction is crucial for real-time eyeblink detection. To this end, we choose to localize the center position of left eye ($P_{left}$) and right eye ($P_{right}$) using off-the-shelf SeetaFace face parsing engine~\cite{Kan2017Funnel} at the first frame. Then, the local eye images are extracted using $P_{left}$ and $P_{right}$ according to Eqn.~\ref{eq:eye_height} and~\ref{eq:eye_width}. Regarding the remaining frames, the local eye images are acquired by tracking the yielded local eye regions of the last frame directly using KCF tracker~\cite{Henriques2014High} due to its high running efficiency. KCF uses the kernelized correlation filter to measure the similarity between 2 signals. And, its discriminative part can be solved within the Discrete Fourier Transform domain to reduce the storage and computation burden by several orders of magnitude. The main technical pipeline for local eye image extraction is shown in Fig.~\ref{figure:local_eye_extraction}.

\begin{table}[t]
  \centering
  \small
  \caption{The declining learning rate that corresponds to the learning step during MS-LSTM training.}
   \begin{tabular}{@{\hspace{5mm}}c@{\hspace{5mm}}c@{\hspace{5mm}}}
    \toprule
    Learning step & Learning rate \\
    \midrule
    1-100 & 0.01 \\
    101-3000 & 0.001 \\
    3001-30000 & 0.0001 \\
    30000-50000 & 0.00001 \\
    \bottomrule
    \end{tabular}%
  \label{table:learning_rate}%
\end{table}%

\section{Implementation details} \label{sec:implementation}

In this section, the essential and important implementation details of the proposed eyeblink detection in the wild approach will be illustrated.

$\bullet$ MS-LSTM is implemented based on the open source machine learning library TensorFlow~\cite{abadi2016tensorflow} for large-scale data processing. It can run on the platforms of CPU, GPU, ASIC and TPU, which is convenient for the developers. Within its architecture, the nodes of a dataflow graph is mapped across many machines in a cluster;

$\bullet$ During the training phase of MS-LSTM, ADAM~\cite{kingma2014adam} is used as the optimizer with the declining learning rate as shown in Table~\ref{table:learning_rate}. The parameters $\beta_1$ and $\beta_2$ in ADAM are set to 0.5 and 0.9 respectively;

$\bullet$ SeetaFace is an open source C++ face parsing engine that can run on CPU with no third-party dependence. Towards our research, it involves 2 key functions, (i.e., face and landmark detection). SeetaFace engine runs in the coarse-to-fine manner, with a novel funnel-structured cascade (FuSt) detection framework. We use its public code with C/C++ programming language at~\url{https://github.com/seetaface/SeetaFaceEngine};

$\bullet$ ADAM is a first-order gradient-based optimization method for stochastic objective functions. It estimates the lower-order moments adaptively, to leverage the optimization performance;

$\bullet$ KCF is implemented using the public code with C/C++ programming language at~\url{https://github.com/vojirt/kcftracker};

$\bullet$ Uniform LBP is implemented by ourselves using C/C++ programming language.

\section{Experiments} \label{sec:experiments}

During experiments to reveal the essential challenges of eyeblink detection in the wild and verify the effectiveness of our proposed eyeblink detection approach, we first compare the performance between our method and the other state-of-the-art eyeblink detection manners~\cite{soukupova2016real,Tabrizi2009Open,Chau2005Real,Morris2002Blink,drutarovsky2014eye} on the proposed HUST-LEBW dataset in Sec.~\ref{sec:experi_blinkdetect}. Since the codes of the approaches employed for comparison are not publicly available and cannot be acquired from the authors, we try our best to implement them by ourselves.

Then to demonstrate the superiority of the proposed MS-LSTM based eyeblink verification approach, we compare it with the other state-of-the-art region-level eyeblink verification methods~\cite{Chau2005Real,Morris2002Blink,drutarovsky2014eye} in Sec.~\ref{sec:experi_blinkverifcation}. To remove the impact of eye location for fair comparison, this test is executed under the assumption that the local eye region has already been successfully extracted in the way of using the manual annotation result directly as depicted in Sec.~\ref{sec:sample_annotation}. Since the approaches in~\cite{soukupova2016real,Tabrizi2009Open} cannot take the local eye image as input, they will not be taken into consideration for comparison in this experimental part.

Consequently, the performance comparison between our eye localization method and the other existing approaches~\cite{soukupova2016real,krolak2012eye,drutarovsky2014eye,Tian2005Real,Tabrizi2009Open} is carried out in Sec.~\ref{sec:experi_eyelocate}. Here, 3 face parsing approaches (i.e., SeetaFace~\cite{Kan2017Funnel}, Intraface~\cite{soukupova2016real}, and MTCNN~\cite{zhang2016joint}) are also compared from the perspectives of both effectiveness and efficiency to justify the reason for why we choose SeetaFace to initially locate the eye center.

The real-time running capacity of our eyeblink detection approach is demonstrated in Sec.~\ref{sec:experi_efficieny}. And, the ablation studies towards MS-LSTM, A-softmax loss function, and low-level eyeblink feature extraction within our method are executed in Sec.~\ref{sec:experi_ms-lstm}, Sec.~\ref{sec:experi_A-softmax} and Sec.~\ref{sec:experi_feature} respectively to reveal the effectiveness of our propositions. The failure cases are given in Sec.~\ref{sec:experi_failures}. And, Sec.~\ref{sec:longvideos} lists the performance of the proposed approach towards the untrimmed video clips.

The experiments run on a laptop with Intel(R) Core(TM) i7-7700HQ CPU @ 2.8GHz (only using one core) and 8 GB RAM memory, under the Windows 10 operation system. During the training phase of MS-LSTM, GPU is used for speed acceleration. But for online test, GPU will not be used.

\begin{table}[t]
  \centering
  \scriptsize
  \caption{Performance comparison among the different eyeblink detection methods on HUST-LEBW dataset. The best performance of each evaluation criteria is shown in boldface. In Tabrizi's method~\cite{Tabrizi2009Open}, eyeblink detection is executed towards left and right eye jointly.}
    \begin{tabular}{@{\hspace{1mm}}c@{\hspace{3mm}}c@{\hspace{3mm}}c@{\hspace{3mm}}c@{\hspace{3mm}}c
    @{\hspace{3mm}}c@{\hspace{1mm}}}
    \toprule
    Method    & Eye idx  & $FR$    & $Recall$ & $Precison$ & $F_1$ score \\
    \midrule
    \multirow{2}[2]{*}{Soukupov{\'a}~\cite{soukupova2016real}} & Left & 0.5820 & 0.3607 & 0.6471 & 0.4632  \\
          & Right & 0.6825 & 0.3016 & 0.5758 & 0.3958  \\
    \midrule
    Tabrizi~\cite{Tabrizi2009Open} & Left$\&$right   & 0.7381 & 0.0714 & 0.4500 & 0.1233  \\
    \midrule
    \multirow{2}[2]{*}{Chau~\cite{Chau2005Real}} & Left & 0.9590 & 0.0164 & \textbf{1.0000} & 0.0323  \\
          & Right & 0.9524 & 0.0000 & 0.0000 & 0.0000  \\
    \midrule
    \multirow{2}[1]{*}{Morris (ver.)~\cite{Morris2002Blink}} & Left & 0.9590 & 0.0164 & 0.6667 & 0.0320  \\
          & Right & 0.9603 & 0.0159 &  \textbf{1.0000} & 0.0313  \\
    \midrule
    \multirow{2}[1]{*}{Morris (hor.)~\cite{Morris2002Blink}} & Left & 0.9590 & 0.0410 & 0.7143 & 0.0775  \\
          & Right & 0.9603 & 0.0238 & 0.7500 & 0.0462  \\
    \midrule
    \multirow{2}[1]{*}{Morris (flow)~\cite{Morris2002Blink}} & Left & 0.9590 & 0.0164 & 0.6667 & 0.0320  \\
          & Right & 0.9603 & 0.0159 & 0.5000 & 0.0308  \\
    \midrule
    \multirow{2}[2]{*}{Drutarovsky~\cite{drutarovsky2014eye}} & Left & 0.7787 & 0.0574 & 0.4118 & 0.1007  \\
          & Right & 0.7857 & 0.0317 & 0.3077 & 0.0576  \\
    \midrule
    \multirow{2}[2]{*}{Our method} & Left & \textbf{0.3197} & \textbf{0.5410} & 0.8919 & \textbf{0.6735} \\
          & Right & \textbf{0.3413} & \textbf{0.4444} & 0.7671 & \textbf{0.5628} \\

    \bottomrule
    \end{tabular}%
   \label{table:blink dtection experiment}%
\end{table}%

\subsection{Performance comparison among the different eyeblink detection methods} \label{sec:experi_blinkdetect}

To evaluate the performance of the different eyeblink detection methods on HUST-LEBW dataset, the criterias of $Recall$, $Precision$ and $F_1$ score are used as below.
\begin{equation}
Recall=\frac{TP}{TP+FN},
\end{equation}
\begin{equation}
Precision=\frac{TP}{TP+FP},
\end{equation}
\begin{equation}
F_1=\frac{2}{\frac{1}{Recall}+\frac{1}{Precision}},
\end{equation}
where $TP$ indicates the number of eyeblink samples recognized correctly; $FN$~\footnote{It is worthy noting that, the eyeblink samples with wrong eye localization result will be regarded as FNs.} and $FP$ denote the number of eyeblink and non-eyeblink samples recognized incorrectly.

Meanwhile, for eyeblink detection in the wild the failure of eye localization essentially weakens the performance. To reveal the impact of this issue, the failure rate ($FR$) of eye localization towards eyeblink samples is given as
\begin{equation}
FR=\frac{N_{miss}+N_{err}}{N_{all}},
\end{equation}
where $N_{miss}$ indicates the number of eyeblink samples that correspond to the case that the eyes cannot be detected at all; $N_{err}$ denotes the number of eyeblink samples that correspond to the case that the eyes  cannot be localized correctly within the all frames; and $N_{all}$ represents the number of eyeblink samples in all. The criteria for judging whether the eye has been correctly localized is given as
\begin{equation}
ME=\frac{MH\left ( \widetilde{P_{loc}},P_{gt}^{loc} \right )}{MH\left ( P^{left}_{gt},P^{right}_{gt} \right )},
\label{eq:me}
\end{equation}
where $MH\left (*,*\right )$ is Manhattan distance function; $ P^{left}_{gt}$ and $P^{right}_{gt}$ indicate the ground-truth position of left and right eye center; $\widetilde{P_{loc}}$ denotes the position of the detected eye center and $P_{gt}^{loc}$ represents its ground-truth position. If $ME>0.4$, we declare that the eye center has not been correctly localized. According to the evaluation criterias above, the comparison among the different eyeblink detection approaches on HUST-LEBW dataset is listed in Table~\ref{table:blink dtection experiment}. It can be observed that:

$\bullet$ Actually, all the eyeblink detection approaches for test (including ours) cannot achieve the satisfactory performance. In summary, their $F_1$ scores cannot exceed 0.7 (0.6735 at most). This phenomenon reveals the fact that, eyeblink detection in the wild is not a trivial but indeed challenging visual recognition task not well solved yet;

$\bullet$ The proposed eyeblink detection approach outperforms the other methods significantly at 3 of the 4 evaluation criteria (except for $Precision$) both on left and right eye, from the perspectives of eye localization and eyeblink verification. That is, the performance gap between our method and the others on $F_1$ score is at least 0.167. This demonstrates the superiority of our proposition towards eyeblink detection in the wild. In some cases, the methods of Chau~\cite{Chau2005Real} and Morris (ver.)~\cite{Morris2002Blink} can yield higher $Precision$ than ours. However, they suffer from low $Recall$ mainly due to high $FR$;

$\bullet$ The challenges of eyeblink detection in the wild essentially derive from the procedures of eye localization and eyeblink verification simultaneously. In particular, all the methods suffers from high $FR$ (over 0.3) on eye localization. Meanwhile, although our approach performs best its $Recall$ and $Precision$ is still relatively low.

\begin{figure}[t]
\centering
\subfigure[Our implementation] {\label{our_res} \includegraphics[height=4.05cm]{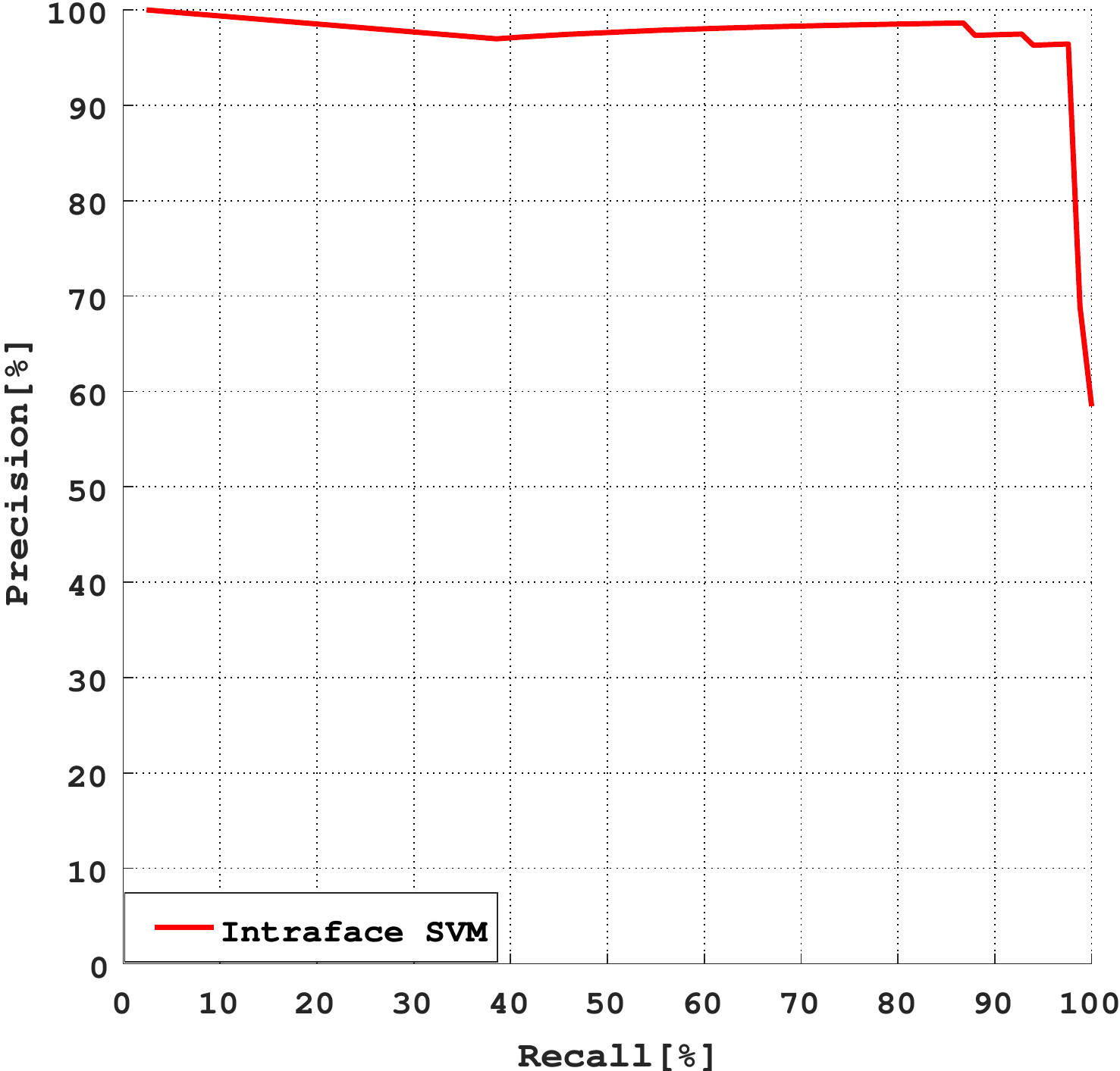}}  
\subfigure[Original results reported in~\cite{soukupova2016real}] {\label{ori_res} \includegraphics[height=4.05cm]{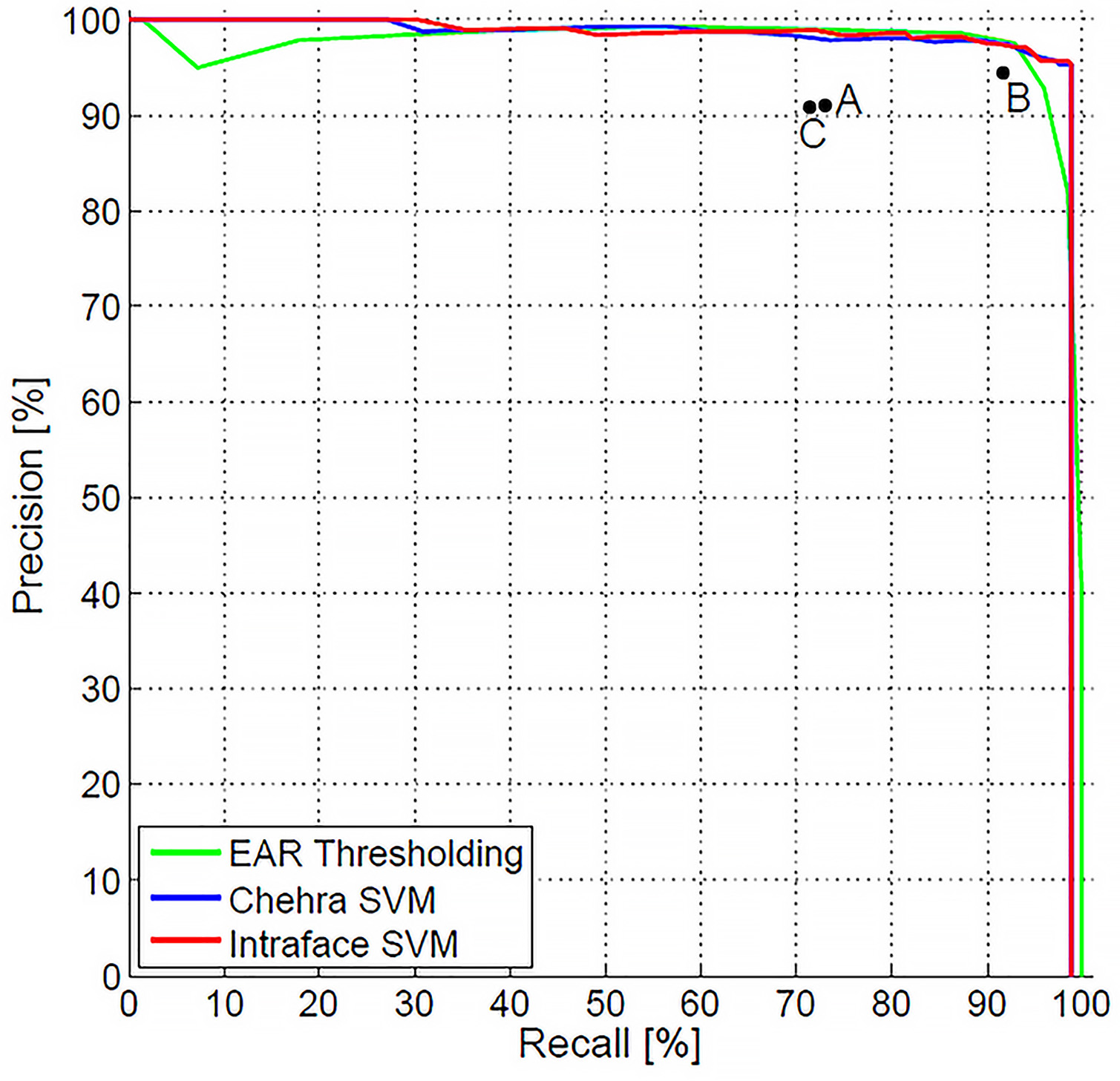}}
\caption{The performance comparison with Precision-Recall curve on ZJU dataset between the method of Soukupov{\'a}~\cite{soukupova2016real} (i.e., Intraface SVM) implemented by us and the original result reported in~\cite{soukupova2016real}. Since the accurate result is not given in~\cite{soukupova2016real}, we choose to cite ``\textbf{Fig. 8(a)}" in ~\cite{soukupova2016real} directly.}
\label{figure:complete}
\end{figure}

Obviously, the approach of Soukupov{\'a}~\cite{soukupova2016real} is our strongest competitor. Since it is implemented by us, the comparison between its original result reported in~\cite{soukupova2016real} and our implementation is conducted on ZJU dataset~\cite{Pan2007Eyeblink} (as shown in Fig.~\ref{figure:complete}) to verify the correctness of our implementation. Particularly, Precision-Recall curve is used as the performance evaluation metric. We can see that, our result is close to the original one in~\cite{soukupova2016real}. This actually reveals the fairness of the conducted experiments. Since the other approaches are much inferior to ours, their implementation correctness will not be verified.

\begin{table}[t]
  \centering
  \scriptsize
  \caption{Performance comparison among the different eyeblink verification methods on HUST-LEBW dataset. The best performance of each evaluation criteria is shown in boldface.}
    \begin{tabular}{@{\hspace{1mm}}c@{\hspace{3mm}}c@{\hspace{3mm}}c@{\hspace{3mm}}c@{\hspace{3mm}}c@{\hspace{1mm}}}
    \toprule
    Method & Eye idx   & $Recall$ & $Precsion$ & $F_1$ score \\
    \midrule
    \multirow{2}[2]{*}{Chau~\cite{Chau2005Real}} & Left & 0.1721  & \textbf{1.0000} & 0.2937  \\
          & Right & 0.2302  & \textbf{0.9656}  & 0.3718  \\
    \midrule
    \multirow{2}[1]{*}{Morris (ver.)~\cite{Morris2002Blink}} & Left & 0.5246  & 0.4741  & 0.4981  \\
          & Right & 0.5635  & 0.5064  & 0.5334  \\
    \midrule
    \multirow{2}[1]{*}{Morris (hor.)~\cite{Morris2002Blink}} & Left & 0.6393  & 0.5342  & 0.5821  \\
          & Right & 0.5476  & 0.5107  & 0.5285  \\
    \midrule
    \multirow{2}[1]{*}{Morris (flow)~\cite{Morris2002Blink}} & Left & 0.4918  & 0.4918  & 0.4918  \\
          & Right & 0.4286  & 0.4741  & 0.4502  \\
    \midrule
    \multirow{2}[1]{*}{Drutarovsky~\cite{drutarovsky2014eye}} & Left & 0.1190  & 0.4757  & 0.1904  \\
          & Right & 0.0952  & 0.2860  & 0.1428  \\
    \midrule
    \multirow{2}[0]{*}{Our method} & Left & \textbf{0.7805 } & 0.7385  & \textbf{0.7589 } \\
          & Right & \textbf{0.8333 } & 0.7778  & \textbf{0.8046 } \\
    \bottomrule
    \end{tabular}%
  \label{table:blink verification experiment}%
\end{table}%

\subsection{Performance comparison among the different eyeblink verification methods} \label{sec:experi_blinkverifcation}

Since the result of eyeblink detection is jointly determined by eye localization and eyeblink verification, to solely verify the superiority of our MS-LSTM based eyeblink verification approach the different methods are compared under the assumption that the local eye region has already been manually extracted in advance. Accordingly, the performance comparison among the different applicable approaches is listed in Table~\ref{table:blink verification experiment}. We can see that:

$\bullet$ Removing the impact of eye localization, the proposed MS-LSTM based eyeblink verification approach still remarkably outperforms the other methods at $F_1$ score by large margins (0.1768 at least), both on left and right eye. This indeed demonstrates the superiority of our proposition over the other manners;

$\bullet$ Even the local eye region has been manually extracted in advance, the performance of the involved approaches is still not promising enough. In particular, the highest $F_1$ score is only 0.8046.  Actually this verifies the fact that eyeblink detection can be regarded as a fine-grained spatial-temporal visual pattern recognition problem of essential challenges, which is also revealed in Fig.~\ref{fig:eyeblink_disreibution} previously;

$\bullet$ Our approach is inferior to Chau's method~\cite{Chau2005Real} at $Precision$. Nevertheless, its $Recall$ and $F_1$ score is much lower than ours.

\subsection{Performance comparison among the different eye localization methods} \label{sec:experi_eyelocate}

\begin{figure}[t]
\centering
\includegraphics[width=0.35\textwidth]{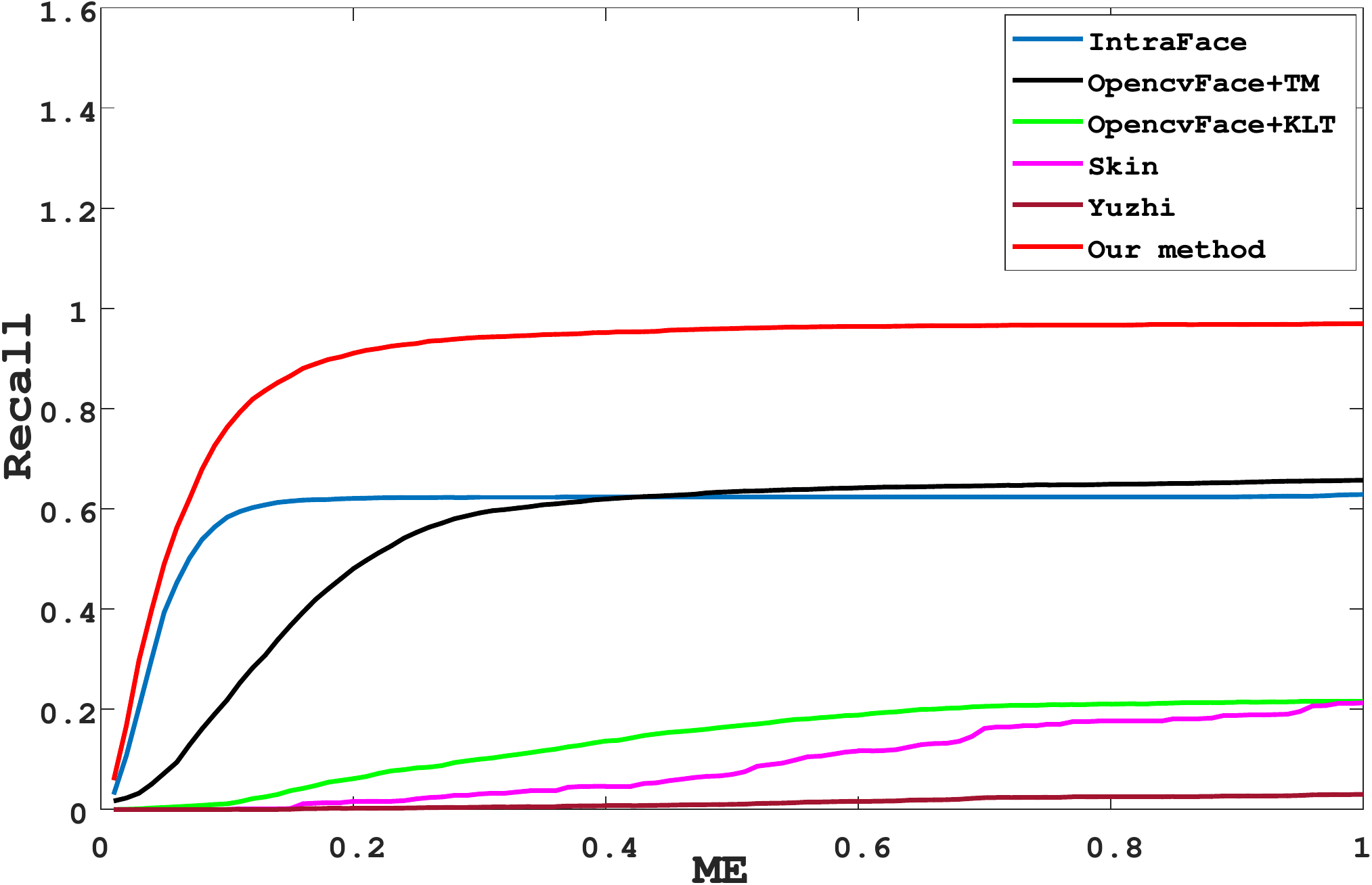}
\caption{The performance comparison among the different eye localization approaches used by the existing eyeblink detection manners.}
\label{figure:eye_location_comp_6}
\end{figure}

Eye localization is the vital step towards most of the eyeblink detection methods. It affects the final performance a lot. Since the existing eyeblink detection approaches generally suffer from high failure rate ($FR$) on eye localization as revealed in Table~\ref{table:blink dtection experiment}, we choose to compare our eye localization approach with the others (i.e., Intraface~\cite{soukupova2016real}, OpencvFace+TM~\cite{krolak2012eye},  OpencvFace+KLT~\cite{drutarovsky2014eye}, Skin~\cite{Tian2005Real} and Yuzhi~\cite{Tabrizi2009Open}) mainly according to $Recall$. The criteria for judging whether the eye has been localized correctly is the same as Sec.~\ref{sec:experi_blinkdetect}, according to $ME$ in Eqn.~\ref{eq:me}. The experiments are executed on all the sample frames within HUST-LEBW dataset. The performance comparison among the different approaches is shown in Fig.~\ref{figure:eye_location_comp_6}. In particular, for compact comparison the average $Recall$ of left and right eye is reported. Obviously our eye localization approach that uses SeetaFace face parsing engine~\cite{Kan2017Funnel} and KCF tracker~\cite{Henriques2014High} is consistently better than the other manners remarkably, corresponding to the different $ME$ thresholds.

\begin{figure}[t]
\centering
\includegraphics[width=0.35\textwidth]{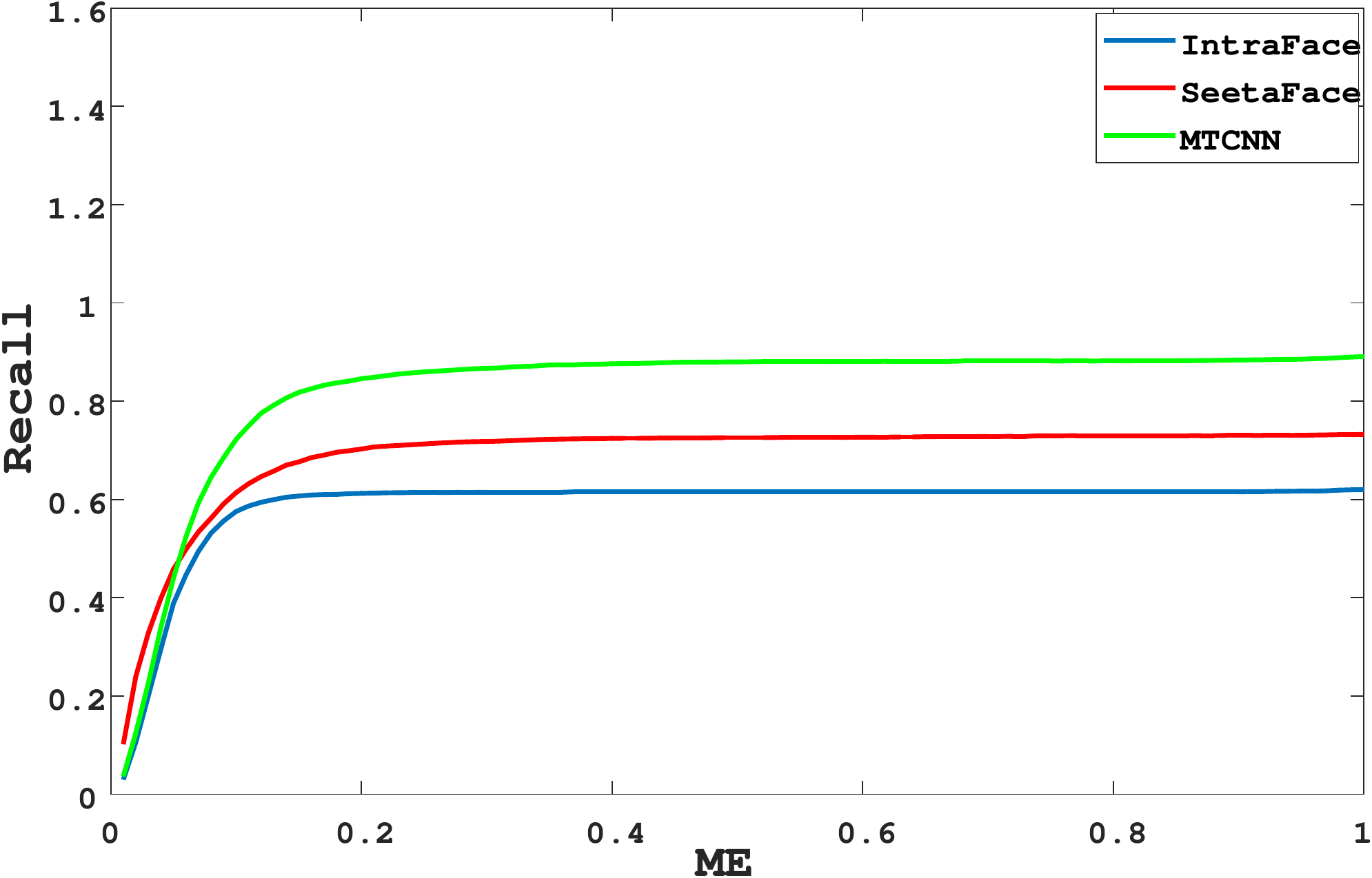}
\caption{The performance comparison among 3 state-of-the-art face parsing methods for eye localization.}
\label{figure:eye_location_comp_3}
\end{figure}

On the other hand, within our approach SeetaFace face parsing engine plays the essential role of localizing eye center initially before tracking. To solely verify its superiority, we compare it with the other 2 state-of-the-art face parsing approaches (i.e., Intraface~\cite{soukupova2016real}, and MTCNN~\cite{zhang2016joint}) from the perspective of effectiveness and efficiency simultaneously. In particular, the performance comparison on effectiveness among the 3 face parsing methods is shown in Fig.~\ref{figure:eye_location_comp_3}. We can see that, in most cases SeetaFace is better than Intraface but inferior to MTCNN. Nevertheless, towards real-time eyeblink detection application running efficiency should also be taken into consideration. We compare the average time consumption of these 3 approaches in Table~\ref{table:average time for locating eye}. It can be observed that, SeetaFace is of the highest running efficiency (i.e., 33.20 ms per frame). Compared to MTCNN, it runs faster of 1 magnitude. Concerning the tradeoff between effectiveness and efficiency for real-time application, we choose SeetaFace as our initial eye localizer.

\begin{table}[t]
  \centering
  \caption{Average time consumption (ms) per frame among the different face parsing approaches for eye localization.}
    \begin{tabular}{cc}
    \toprule
    Method & Time consumption \\
    \midrule
    SeetaFace~\cite{Kan2017Funnel} & \textbf{33.20} \\
    Intraface~\cite{soukupova2016real} & 85.89 \\
    MTCNN~\cite{zhang2016joint} & 503.07 \\
    \bottomrule
    \end{tabular}%
  \label{table:average time for locating eye}%
\end{table}%

\subsection{Real-time online running capacity verification} \label{sec:experi_efficieny}

In this subsection, we will verify that our proposed eyeblink detection method is of real-time online running capacity on a personal computer with Intel(R) Core(TM) i7-7700HQ CPU @ 2.8GHz (only using one core). The average online running time consumption per frame of the main procedures within our method is listed in Table~\ref{table:time consumation of MSLTM}. It can be observed that, the main time consumption is costed by SeetaFace engine for initial eye localization with 33.20 ms. However, it will be executed only on the first frame towards an eyeblink sample. And, the procedures of eye tracking, eyeblink feature extraction, and eyeblink verification are extremely fast with the time consumption of only 7.87 ms in all. We can make a summary that, the initial eye localization procedure can run with the speed over 29 FPS. When turning to eye tracking phase, the proposed eyeblink detection method can run with the speed over 127 FPS. Overall, our approach meets the real-time running requirement (i.e., with the speed over 25 FPS).

\begin{table}[t]
  \centering
  \footnotesize
  \caption{The average online running time consumption (ms) per frame of the main procedures within the proposed eyeblink detection approach.}
        \begin{tabular}{cc}
    \toprule
    Procedure & Time consumption \\
    \midrule
    Initial eye localization (SeetaFace) & 33.20 \\
    Eye tracking (KCF) & 6.06 \\
   \midrule
  Eyeblink feature extraction (uniform LBP) & 0.32 \\
  \midrule
  Eyeblink verification (MS-LSTM) & 1.49 \\

    \bottomrule
    \end{tabular}%
 \label{table:time consumation of MSLTM}%
\end{table}

\subsection{Ablation study 1: MS-LSTM} \label{sec:experi_ms-lstm}

MS-LSTM is proposed by us to address the problem of eyeblink verification. From the network structure perspective, it holds 2 main modifications compared with the original LSTM model to alleviate the multiple temporal scale problem within eyeblink. One is to stack multiple LSTM layers. And, the other is to involve multiple temporal scale feature. Here, we will verify the effectiveness of the 2 modifications respectively. The experiments are executed under the assumption that the local eye region has already been manually extracted in advance, which is the same as Sec~\ref{sec:experi_blinkverifcation}.

\begin{table}[t]
  \centering
  \scriptsize
  \caption{Performance comparison among MS-LSTMs with the different numbers of stacked LSTM layers.}
    \begin{tabular}{ccccc}
    \toprule
    Eye idx   & Layer number & $Recal$ & $Precision$ & $F_1$ score \\
    \midrule
    \multirow{4}[2]{*}{Left} & 1     & 0.6098  & \textbf{0.8929} & 0.7246  \\
          & 2     & \textbf{0.7805} & 0.7385  & 0.7589  \\
          & 3     & 0.6992  & 0.8350  & \textbf{0.7611} \\
          & 4     & 0.7073  & 0.8056  & 0.7532  \\
    \midrule
    \multirow{4}[2]{*}{Right} & 1     & 0.7619  & 0.7934  & 0.7773  \\
          & 2     & \textbf{0.8333} & 0.7778  & \textbf{0.8046} \\
          & 3     & 0.7857  & 0.7984  & 0.7920  \\
          & 4     & 0.7629  & \textbf{0.8276} & 0.7934  \\
    \midrule
    \multirow{4}[1]{*}{Average} & 1     & 0.6859  & \textbf{0.8432} & 0.7510  \\
          & 2     & \textbf{0.8069} & 0.7582  & \textbf{0.7818} \\
          & 3     & 0.7425  & 0.8167  & 0.7766  \\
          & 4     & 0.7351  & 0.8166  & 0.7733  \\
    \bottomrule
    \end{tabular}%
    \label{table:LSTM_layer}%
\end{table}%

\textbf{Stack multiple LSTM layers.} The number of the stacked LSTM layers is set from 1 to 4. The performance comparison among them is listed in Table~\ref{table:LSTM_layer}. It can be seen that:

$\bullet$ Compared to the original LSTM model with only 1 layer, adding the layer number can consistently leverage the performance on $Recall$ and $F_1$ score in all the test cases. However, it may weaken $Precision$. Overall, stacking multiple LSTM layers is an effective way to enhance eyeblink verification result comprehensively.

$\bullet$ Setting the layer number to 2 can achieve the best average performance on $Recall$ and $F_1$ score. Accordingly, the layer number within the proposed MS-LSTM model is empirically set to 2 for eyeblink verification.

 \begin{table}[t]
  \centering
  \scriptsize
  \caption{Performance comparison among MS-LSTMs with the different temporal scale numbers.}
    \begin{tabular}{ccccc}
    \toprule
    Eye idx   & Scale number   & $Recall$ & $Precision$ & $F_1$ score \\
    \midrule
    \multirow{5}[2]{*}{Left} & 1     & \textbf{0.8455} & 0.7123  & \textbf{0.7732} \\
          & 2     & 0.7805  & 0.7385  & 0.7589  \\
          & 3     & 0.6585  & \textbf{0.9000} & 0.7606  \\
          & 4     & 0.6016  & 0.8916  & 0.7184  \\
          & 5     & 0.7480  & 0.7667  & 0.7572  \\
    \midrule
    \multirow{5}[2]{*}{Right} & 1     & 0.5952  & \textbf{0.8824} & 0.7109  \\
          & 2     & \textbf{0.8333} & 0.7778  & \textbf{0.8046} \\
          & 3     & 0.7460  & 0.7833  & 0.7642  \\
          & 4     & 0.7619  & 0.7742  & 0.7680  \\
          & 5     & 0.7302  & 0.7863  & 0.7572  \\
    \midrule
    \multirow{5}[2]{*}{Average} & 1     & 0.7204  & 0.7974  & 0.7421  \\
          & 2     & \textbf{0.8069} & 0.7582  & \textbf{0.7818} \\
          & 3     & 0.7023  & \textbf{0.8417} & 0.7624  \\
          & 4     & 0.6818  & 0.8329  & 0.7432  \\
          & 5     & 0.7391  & 0.7765  & 0.7572  \\
    \bottomrule
    \end{tabular}%
  \label{table:Temporal_scale}%
\end{table}%

\textbf{Multiple temporal scale feature.} The temporal scale number is set from 1 to 5. The performance comparison among them is listed in Table~\ref{table:Temporal_scale}. We can see that:

$\bullet$ Involving multiple temporal scale feature essentially leverages the performance of eyeblink verification, especially from the perspectives of average $Recall$, $Precision$ and $F_1$ score. This actually demonstrates the effectiveness of our proposition on extracting multiple temporal scale feature for eyeblink characterization within MS-LSTM model;

$\bullet$ Setting the temporal scale number to 2 can achieve the best average performance on $Recall$ and $F_1$ score. Accordingly, the temporal scale number of the proposed MS-LSTM model is empirically set to 2 for eyeblink verification.

\begin{table}[t]
  \centering
  \scriptsize
  \caption{Performance comparison between softmax and A-softmax loss function for eyeblink verification.}
    \begin{tabular}{@{\hspace{1.5mm}}c@{\hspace{1.5mm}}c@{\hspace{1.5mm}}c@{\hspace{1.5mm}}c@{\hspace{1.5mm}}c@{\hspace{1.5mm}}}
    \toprule
    Eye idx   & Loss function & $Recall$ & $Precision$ & $F_1$ score \\
    \midrule
    \multirow{2}[2]{*}{Left} & Softmax & 0.7497  & 0.7304  & 0.7394  \\
          & A-softmax & \textbf{0.7805}  & \textbf{0.7385}  & \textbf{0.7589}  \\
    \midrule
    \multirow{2}[2]{*}{Right} & Softmax & 0.6726  & 0.7581  & 0.7128  \\
          & A-softmax & \textbf{0.8333}  & \textbf{0.7778}  & \textbf{0.8046}  \\
    \midrule
    \multirow{2}[2]{*}{Average} & Softmax & 0.7112  & 0.7443  & 0.7261  \\
          & A-softmax & \textbf{0.8069}  & \textbf{0.7582}  & \textbf{0.7818}  \\
    \bottomrule
    \end{tabular}%
  \label{tab:fine_grained}%
\end{table}

\begin{table*}[t]
  \centering
  \scriptsize
  \caption{The performance comparison among the different visual descriptors under the appearance-motion eyeblink feature extraction mechanism. In particular, ``app." indicates appearance feature and ``motion" denotes motion feature for eyeblink characterization.}
    \begin{tabular}{cc|ccc|ccc|rrr}
    \toprule
    \multirow{2}[2]{*}{Descriptor} & \multirow{2}[2]{*}{Mechanism} & \multicolumn{3}{c|}{Left eye} & \multicolumn{3}{c|}{Right eye} & \multicolumn{3}{c}{Average} \\
          &       & $Recall$ & $Precision$ & $F_1$ score & $Recall$ & $Precision$ & $F_1$ score & \multicolumn{1}{c}{$Recall$} & \multicolumn{1}{c}{$Precision$} & \multicolumn{1}{c}{$F_1$ score} \\
    \midrule
    \multirow{3}[2]{*}{Uniform LBP~\cite{ahonen2006face} } & App.   & 0.7398  & \textbf{0.7459} & 0.7429  & 0.7857  & 0.7444  & 0.7645  & 0.7628  & 0.7452  & 0.7537  \\
          & Motion & 0.7925  & 0.5250  & 0.6316  & 0.6667  & 0.6389  & 0.6525  & 0.7296  & 0.5820  & 0.6421  \\
    & App.+motion & 0.7805  & 0.7385  & \textbf{0.7589} & 0.8333  & 0.7778  & 0.8046  & 0.8069  & 0.7582  & \textbf{0.7818} \\

    \midrule
    \multirow{3}[2]{*}{HOG~\cite{Dalal2005Histograms}} & App.   & 0.6911  & 0.5944  & 0.6391  & 0.8175  & 0.6242  & 0.7079  & 0.7543  & 0.6093  & 0.6735  \\
          & Motion & 0.7698  & 0.5834  & 0.6644 & 0.8182  & 0.5934  & 0.6879 & 0.7940  & 0.5884  & 0.6762  \\
    & App.+motion & 0.7398  & 0.7054  & 0.7222 & 0.8016  & \textbf{0.8347} & \textbf{0.8178} & 0.7707  & \textbf{0.7701} & 0.7700  \\

    \midrule
    \multirow{3}[2]{*}{Haar~\cite{Liu2012}} & App.  & 0.8115  & 0.5824  & 0.6781  & 0.6667  & 0.6512  & 0.6588  & 0.7391  & 0.6168  & 0.6685  \\
          & Motion & 0.6395  & 0.6763  & 0.6573  & 0.6561  & 0.7007  & 0.6776  & 0.6478  & 0.6885  & 0.6675  \\
    & App.+motion & \textbf{0.8130} & 0.5848  & 0.6803  & \textbf{0.8413} & 0.6463  & 0.7310  & \textbf{0.8272} & 0.6156  & 0.7057  \\

    \bottomrule
    \end{tabular}%
\label{table:low-level_feature}%
\end{table*}%

\subsection{Ablation study 2: A-softmax loss function} \label{sec:experi_A-softmax}

As revealed in Fig.~\ref{fig:eyeblink_disreibution}, eyeblink verification can be regarded as a fine-grained binary spatial-temporal pattern recognition problem. To ensure the classification margin between eyeblink and non-eyeblink classes, A-softmax loss function is used within MS-LSTM model. To verify its superiority, we compare it with the original softmax loss function. The experiments are executed under the assumption that the local eye region has already been manually extracted in advance, which is the same as Sec~\ref{sec:experi_blinkverifcation}. The performance comparison between these 2 loss functions is listed in Table~\ref{tab:fine_grained}. It is impressive that A-softmax loss function consistently outperforms the original softmax loss function in all test cases, especially on the average $Recall$ and $F_1$ score. This indeed demonstrates the effectiveness of our proposition that applies A-softmax loss function to address eyeblink verification.

\subsection{Ablation study 3: low-level eyeblink feature extraction} \label{sec:experi_feature}

To effectively characterize eyeblink, we propose to extract low-level appearance and motion feature simultaneously as the input of MS-LSTM using uniform LBP. To justify the superiority of our low-level eyeblink feature extraction method, we conduct experiments in 2 folders. First, uniform LBP is compared with the other 2 well-established visual descriptors (i.e., HOG~\cite{Dalal2005Histograms} and Haar~\cite{Liu2012}). Meanwhile, the effectiveness of the mechanism on extracting appearance and motion feature simultaneously for eyeblink characterization is also verified. The experiments are executed under the assumption that the local eye region has already been manually extracted in advance, which is the same as Sec~\ref{sec:experi_blinkverifcation}. The comprehensive performance comparison is listed in Table~\ref{table:low-level_feature}. It can be observed that:

$\bullet$ Among the 3 visual descriptors for test, uniform LBP can achieve the best result on the average $F_1$ score. Its performance on the average $Recall$ and $Precision$ is also comparable to the best one. Overall, uniform LBP is the optimal choice for eyeblink detection;

$\bullet$ For all the 3 visual descriptors the mechanism of extracting appearance and motion feature simultaneously can essentially leverage the performance in most cases, compared to using only one type feature.

$\bullet$ In addition, the running time comparison of the 3 visual descriptors is listed in Table~\ref{table:average_time_descriptor}. We can see that, uniform LBP is of the fastest running speed.

The experimental results above indeed demonstrate the effectiveness of our proposed low-level eyeblink feature extraction approach.

\subsection{Failure cases of eyeblink detection in the wild} \label{sec:experi_failures}

\begin{figure}[t]
\centering
\subfigure[Face detection failure] {\label{fig:face_failure} \includegraphics[width=0.45\textwidth]{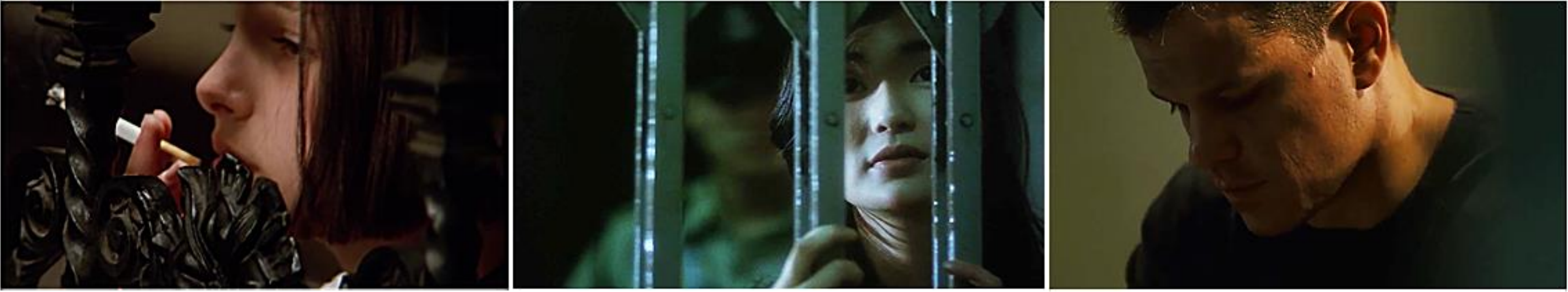}}  
\subfigure[Initial eye localization failure] {\label{fig:eye_location_falure} \includegraphics[width=0.45\textwidth]{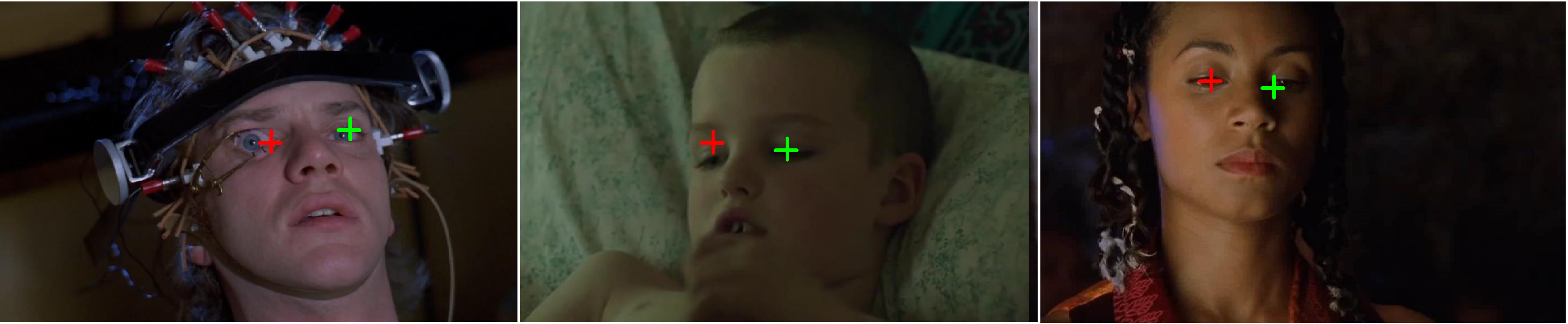}}
\subfigure[Eye tracking failure] {\label{fig:tracking_falure} \includegraphics[width=0.45\textwidth]{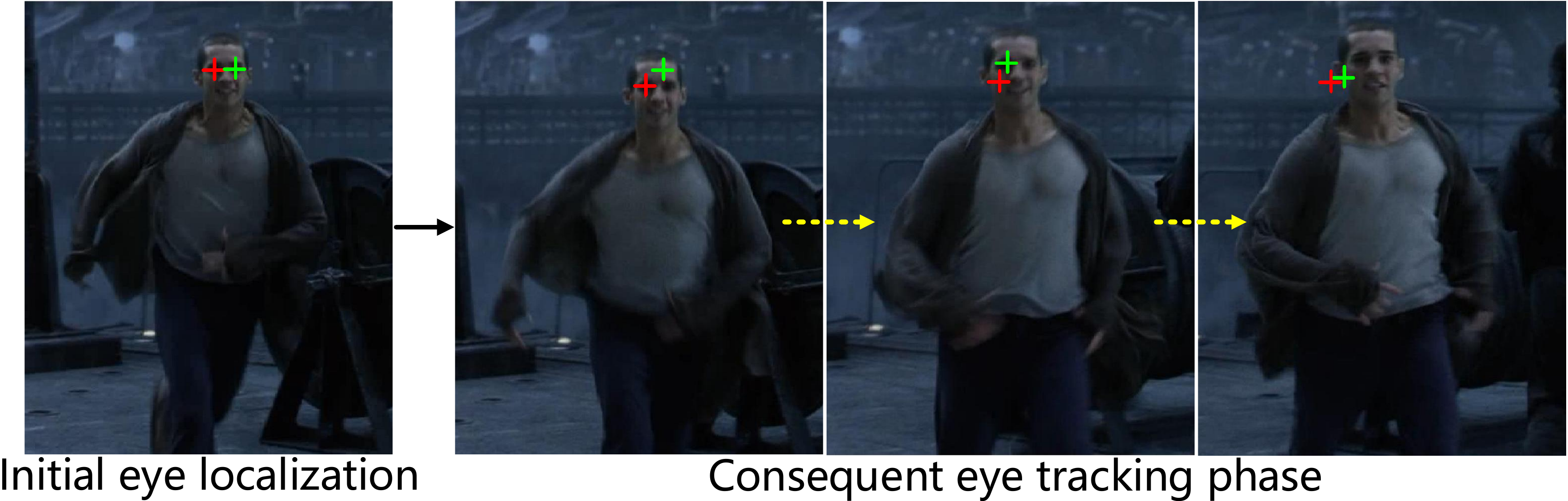}}
\subfigure[False positive non-eyeblink sample] {\label{fig:false_positive}
\includegraphics[width=0.45\textwidth]{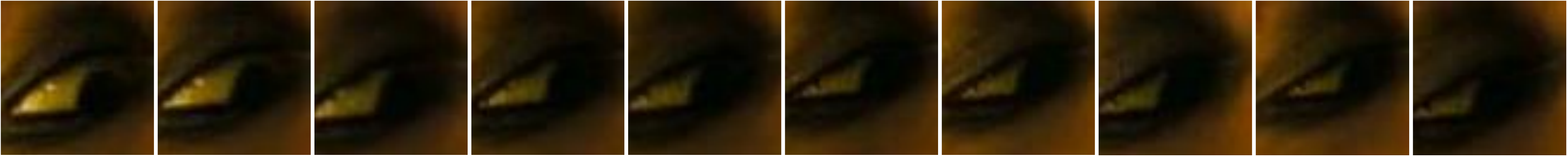}}
\subfigure[False negative eyeblink sample] {\label{fig:false_negative}
\includegraphics[width=0.45\textwidth]{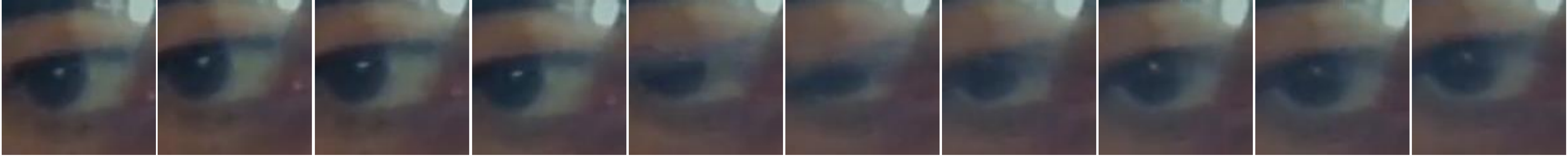}}
\caption{The failure cases towards eyeblink detection in the wild of our proposed approach. In particular,  ``$\textcolor{red}{+}$" indicates the position of right eye and ``$\textcolor{green}{+}$" denotes the position of left eye.}
\label{fig:eyeblink_failure_case}
\end{figure}

From Sec.~\ref{sec:experi_blinkdetect} to Sec.~\ref{sec:experi_feature}, quantitative performance evaluation is executed to demonstrate its effectiveness and superiority of our proposition. Here, qualitative analysis will be conducted to show the defects of our proposition towards in the wild application scenario. Accordingly the intuitive failure case examples are given in Fig.~\ref{fig:eyeblink_failure_case} from different perspectives, aiming to reveal some insights towards eyeblink detection in the wild and indicate the future research avenue. We can see that accurate face detection, eye localization and eye tracking is still remaining as the challenging visual tasks under the unconstrained ``in the wild" conditions, although numerous efforts have already been paid. The challenges actually derive from the dramatic variation on human attribute, human pose, illumination, and scene conditions. From Fig.~\ref{fig:tracking_falure}, the fast movement of human is also a critical issue to impair eye tracking. Meanwhile, the makeup on eye may also confuse the classifier during the phase of eyeblink verification as shown in Fig.~\ref{fig:false_positive}. What is more challenging is that within some eyeblink samples the eyes are not fully closed as shown in Fig.~\ref{fig:false_negative}, which may be caused by the relatively low frame rate of camera. These require us to extract more discriminative spatial-temporal feature for eyeblink characterization.

\begin{table}[t]
  \centering
  \scriptsize
  \caption{Average time consumption (ms) per frame among the different visual descriptors for eyeblink characterization.}
    \begin{tabular}{cc}
    \toprule
    Descriptor & Time consumption(ms) \\
    \midrule
    Uniform LBP~\cite{ahonen2006face} & \textbf{0.322} \\
    HOG~\cite{Dalal2005Histograms} & 0.344 \\
    Haar~\cite{Liu2012} & 0.650 \\
    \bottomrule
    \end{tabular}%
  \label{table:average_time_descriptor}%
\end{table}%

\subsection{Eyeblink detection towards untrimmed video clip} \label{sec:longvideos}

Here our proposed eyeblink detection approach is tested on the untrimmed video clips, which is more close to the practical applications. To this end, a sub-dataset of HUST-LEBW is built. In particular, 90 untrimmed video clips of the average length of 51.30 frames are captured to cover all the 127 raw eyeblink samples within HUST-LEBW for test. A 10-frame sliding temporal window is set to decide the start and end position of eyeblink, with the stride of 1 frame. Temporal NMS is executed to reduce the redundant predictions, with the IoU threshold~\cite{shou2016temporal} of $0.33$ and eyeblink confidence score threshold of 0.5. Performance evaluation is executed according to THUMOS-overlap 0.5 criterion~\cite{THUMOS14}, and average precision (AP)~\cite{THUMOS14} is reported. During online test, eye localization using SeetaFace engine will be re-executed when the eye tracking score of KCF is below 0.25. The test results are listed in Table~\ref{tab:longvideo}. We can see that, the performance of the proposed eyeblink detection approach is actually not satisfactory enough with the relatively low AP and high FR on eye localization. This essentially verifies the challenges of eyeblink detection in the wild, especially towards the practical applications. However, the running speed of our method is still nearly real-time (27.55 FPS in average).

\begin{table}[t]
  \centering
  \scriptsize
  \caption{Performance of eyeblink detection on untrimmed video clips. }
    \begin{tabular}{cccc}
    \toprule
        Eye idx  & AP  & FR  & Average time (ms) \\
    \midrule
    Left eye & 0.2942 & 0.1633 & 36.33 \\
    Right eye & 0.3185  &  0.1531  & 36.33 \\
    \bottomrule
    \end{tabular}%
  \label{tab:longvideo}%
\end{table}%

\section{Conclusions} \label{sec:conclusions}

In this work, we shed the light to the research field of eyeblink detection in the wild not well studied before. First, an eyeblink detection in the wild dataset (HUST-LEBW) is built. Second, MS-LSTM model is proposed to address the fine-grained spatial-temporal pattern recognition problem within eyeblink detection. Third, an effective and efficient eyeblink feature extraction approach is proposed to capture appearance and motion information simultaneously. Meanwhile, our eyeblink detection method can run in real-time on a normal laptop without using parallel computing. The extensive experiments verify the challenges of eyeblink detection in the wild, and demonstrate the superiority of the proposed approach.

\section*{Acknowledgment}
This work is jointly supported by the National Key R\&D Program of China (No. 2018YFB1004600), National Natural Science Foundation of China (Grant No. 61502187, 61702182, 61876211 and 61602193), Equipment Pre-research Field Fund of China (Grant No. 61403120405),  start-up funds from University at Buffalo, Fundamental Research Funds for the Central Universities (Grant No. 2019kfyXKJC024), Hunan Provincial Natural Science Foundation of China (Grant 2018JJ3254) and National Key Laboratory Open Fund of China (Grant No. 6142113180211).

Joey Tianyi Zhou is supported by Programmatic Grant No. A18A1b0045 from the Singapore government's Research, Innovation and Enterprise 2020 plan (Advanced Manufacturing and Engineering domain).

\ifCLASSOPTIONcaptionsoff
\newpage
\fi

\bibliographystyle{IEEEtran}
\bibliography{paper}

\begin{thebibliography}{10}
\providecommand{\url}[1]{#1}
\csname url@samestyle\endcsname
\providecommand{\newblock}{\relax}
\providecommand{\bibinfo}[2]{#2}
\providecommand{\BIBentrySTDinterwordspacing}{\spaceskip=0pt\relax}
\providecommand{\BIBentryALTinterwordstretchfactor}{4}
\providecommand{\BIBentryALTinterwordspacing}{\spaceskip=\fontdimen2\font plus
\BIBentryALTinterwordstretchfactor\fontdimen3\font minus
  \fontdimen4\font\relax}
\providecommand{\BIBforeignlanguage}[2]{{%
\expandafter\ifx\csname l@#1\endcsname\relax
\typeout{** WARNING: IEEEtran.bst: No hyphenation pattern has been}%
\typeout{** loaded for the language `#1'. Using the pattern for}%
\typeout{** the default language instead.}%
\else
\language=\csname l@#1\endcsname
\fi
#2}}
\providecommand{\BIBdecl}{\relax}
\BIBdecl

\bibitem{Perelman2014Detecting}
B.~S. Perelman, ``Detecting deception via eyeblink frequency modulation,''
  \emph{Peerj}, vol.~2, no.~2, p. e260, 2014.

\bibitem{Bergasa2006Real}
L.~M. Bergasa, J.~Nuevo, M.~A. Sotelo, R.~Barea, and M.~E. Lopez, ``Real-time
  system for monitoring driver vigilance,'' \emph{IEEE Trans. on Intelligent
  Transportation Systems}, vol.~7, no.~1, pp. 63--77, 2006.

\bibitem{Pan2007Eyeblink}
G.~Pan, L.~Sun, Z.~Wu, and S.~Lao, ``Eyeblink-based anti-spoofing in face
  recognition from a generic webcamera,'' in \emph{Proc. IEEE International
  Conference on Computer Vision (ICCV)}, 2007, pp. 1--8.

\bibitem{Rosenfield2011Computer}
M.~Rosenfield, ``Computer vision syndrome: a review of ocular causes and
  potential treatments.'' \emph{Ophthalmic \& Physiological Optics}, vol.~31,
  no.~5, pp. 502--515, 2011.

\bibitem{Ji2014Real}
Q.~Ji and X.~Yang, ``Real-time eye, gaze, and face pose tracking for monitoring
  driver vigilance,'' \emph{Real-Time Imaging}, vol.~8, no.~5, pp. 357--377,
  2014.

\bibitem{Wu2008Development}
J.~D. Wu and T.~R. Chen, ``Development of a drowsiness warning system based on
  the fuzzy logic images analysis,'' \emph{Expert Systems with Applications},
  vol.~34, no.~2, pp. 1556--1561, 2008.

\bibitem{Dong2008Driver}
W.~Dong, P.~Qu, and J.~Han, ``Driver fatigue detection based on fuzzy fusion,''
  in \emph{Proc. Chinese Control and Decision Conference (CCDC)}, 2008, pp.
  2640--2643.

\bibitem{Tabrizi2009Open}
P.~R. Tabrizi and R.~A. Zoroofi, ``Open/closed eye analysis for drowsiness
  detection,'' in \emph{Proc. Image Processing Theory, Tools and Applications
  Workshop (IPTAW)}.\hskip 1em plus 0.5em minus 0.4em\relax IEEE, 2008, pp.
  1--7.

\bibitem{krolak2012eye}
A.~Kr{\'o}lak and P.~Strumi{\l}{\l}o, ``Eye-blink detection system for
  human--computer interaction,'' \emph{Universal Access in the Information
  Society}, vol.~11, no.~4, pp. 409--419, 2012.

\bibitem{Chau2005Real}
M.~Chau and M.~Betke, ``Real time eye tracking and blink detection with usb
  cameras,'' \emph{Cas Computer Science Technical Reports}, 2005.

\bibitem{Hong2008Drivers}
T.~Hong and H.~Qin, ``Drivers drowsiness detection in embedded system,'' in
  \emph{Proc. IEEE International Conference on Vehicular Electronics and Safety
  (ICVES)}, 2008, pp. 1--5.

\bibitem{Morris2002Blink}
T.~Morris, P.~Blenkhorn, and F.~Zaidi, ``Blink detection for real-time eye
  tracking,'' \emph{Journal of Network \& Computer Applications}, vol.~25,
  no.~2, pp. 129--143, 2002.

\bibitem{drutarovsky2014eye}
T.~Drutarovsky and A.~Fogelton, ``Eye blink detection using variance of motion
  vectors,'' in \emph{Proc. European Conference on Computer Vision Workshop
  (ECCVW)}.\hskip 1em plus 0.5em minus 0.4em\relax Springer, 2014, pp.
  436--448.

\bibitem{talkingface}
``Talking face video,'' \url{
  http://www-prima.inrialpes.fr/FGnet/data/01-TalkingFace/talking_face.html},
  {F}ace\&{G}esture {R}ecognition {W}orking {G}roup, {IST}-2000-26434.

\bibitem{radlak2015silesian}
K.~Radlak, M.~Bozek, and B.~Smolka, ``Silesian deception database: Presentation
  and analysis,'' in \emph{Proc. ACM Multimodal Deception Detection Workshop
  (MDDW)}, 2015, pp. 29--35.

\bibitem{researer}
``Researchers' night,''
  \url{https://ec.europa.eu/research/mariecurieactions/actions/european-researchers-night_en}.

\bibitem{Kan2017Funnel}
M.~Kan, M.~Kan, S.~Shan, S.~Shan, and X.~Chen, ``Funnel-structured cascade for
  multi-view face detection with alignment-awareness,'' \emph{Neurocomputing},
  vol. 221, no.~C, pp. 138--145, 2017.

\bibitem{Henriques2014High}
J.~F. Henriques, C.~Rui, P.~Martins, and J.~Batista, ``High-speed tracking with
  kernelized correlation filters,'' \emph{IEEE Trans. on Pattern Analysis \&
  Machine Intelligence}, vol.~37, no.~3, pp. 583--596, 2014.

\bibitem{ahonen2006face}
T.~Ahonen, A.~Hadid, and M.~Pietikainen, ``Face description with local binary
  patterns: Application to face recognition,'' \emph{IEEE Trans. on Pattern
  Analysis \& Machine Intelligence}, no.~12, pp. 2037--2041, 2006.

\bibitem{Lee2010Blink}
W.~O. Lee, E.~C. Lee, and R.~P. Kang, ``Blink detection robust to various
  facial poses,'' \emph{Journal of Neuroscience Methods}, vol. 193, no.~2, p.
  356, 2010.

\bibitem{torricelli2009adaptive}
D.~Torricelli, M.~Goffredo, S.~Conforto, and M.~Schmid, ``An adaptive blink
  detector to initialize and update a view-basedremote eye gaze tracking system
  in a natural scenario,'' \emph{Pattern Recognition Letters}, vol.~30, no.~12,
  pp. 1144--1150, 2009.

\bibitem{soukupova2016real}
T.~Soukupov{\'a} and J.~Cech, ``Real-time eye blink detection using facial
  landmarks,'' in \emph{Proc. Computer Vision Winter Workshop (CVWW)}, 2016.

\bibitem{Sun2009Robust}
R.~Sun and Z.~Ma, ``Robust and efficient eye location and its state
  detection,'' in \emph{Proc. International Symposium on Advances in
  Computation and Intelligence (ISACI)}, 2009, pp. 318--326.

\bibitem{Liu2012}
Z.~Liu and H.~Ai, ``Automatic eye state recognition and closed-eye photo
  correction,'' in \emph{Porc. International Conference on Pattern Recognition
  (ICPR)}, 2012, pp. 1--4.

\bibitem{Dalal2005Histograms}
N.~Dalal and B.~Triggs, ``Histograms of oriented gradients for human
  detection,'' in \emph{Proc. IEEE Conference on Computer Vision and Pattern
  Recognition (CVPR)}, 2005, pp. 886--893.

\bibitem{Tan2006Detecting}
H.~Tan and Y.~J. Zhang, ``Detecting eye blink states by tracking iris and
  eyelids,'' \emph{Pattern Recognition Letters}, vol.~27, no.~6, pp. 667--675,
  2006.

\bibitem{Bradski2000The}
G.~Bradski and A.~Kaehler, ``Opencv,'' \emph{Dr. Dobb’s Journal of Software
  Tools}, vol.~3, 2000.

\bibitem{DBLP:journals/corr/YinL17}
\BIBentryALTinterwordspacing
X.~Yin and X.~Liu, ``Multi-task convolutional neural network for face
  recognition,'' \emph{CoRR}, vol. abs/1702.04710, 2017. [Online]. Available:
  \url{http://arxiv.org/abs/1702.04710}
\BIBentrySTDinterwordspacing

\bibitem{zhang2016joint}
K.~Zhang, Z.~Zhang, Z.~Li, and Y.~Qiao, ``Joint face detection and alignment
  using multitask cascaded convolutional networks,'' \emph{IEEE Signal
  Processing Letters}, vol.~23, no.~10, pp. 1499--1503, 2016.

\bibitem{cao2017early}
H.~Cao, K.~Zhou, X.~Chen, and X.~Zhang, ``Early chatter detection in end
  milling based on multi-feature fusion and 3$\sigma$ criterion,'' \emph{The
  International Journal of Advanced Manufacturing Technology}, vol.~92, no.
  9-12, pp. 4387--4397, 2017.

\bibitem{stigler1986history}
S.~M. Stigler, \emph{The history of statistics: The measurement of uncertainty
  before 1900}.\hskip 1em plus 0.5em minus 0.4em\relax Harvard University
  Press, 1986.

\bibitem{oguz1996proportion}
{\"O}.~Oguz, ``The proportion of the face in younger adults using the thumb
  rule of leonardo da vinci,'' \emph{Surgical and Radiologic Anatomy}, vol.~18,
  no.~2, pp. 111--114, 1996.

\bibitem{maaten2008visualizing}
L.~V. Der~Maaten and G.~E. Hinton, ``Visualizing data using t-sne,''
  \emph{Journal of Machine Learning Research}, vol.~9, pp. 2579--2605, 2008.

\bibitem{hochreiter1997long}
S.~Hochreiter and J.~Schmidhuber, ``Long short-term memory,'' \emph{Neural
  computation}, vol.~9, no.~8, pp. 1735--1780, 1997.

\bibitem{liu2017global}
J.~Liu, G.~Wang, P.~Hu, L.-Y. Duan, and A.~C. Kot, ``Global context-aware
  attention lstm networks for 3d action recognition,'' in \emph{Proc. IEEE
  Conference on Computer Vision and Pattern Recognition (CVPR)}, vol.~7, 2017,
  p.~43.

\bibitem{hopfield1982neural}
J.~J. Hopfield, ``Neural networks and physical systems with emergent collective
  computational abilities,'' \emph{Proc. of the National Academy of Sciences
  (NAS)}, vol.~79, no.~8, pp. 2554--2558, 1982.

\bibitem{hermans2013training}
M.~Hermans and B.~Schrauwen, ``Training and analysing deep recurrent neural
  networks,'' in \emph{Proc. Advances in Neural Information Processing Systems
  (NIPS)}, 2013, pp. 190--198.

\bibitem{zhu2016co-occurrence}
W.~Zhu, C.~Lan, J.~Xing, W.~Zeng, Y.~Li, L.~Shen, and X.~Xie, ``Co-occurrence
  feature learning for skeleton based action recognition using regularized deep
  lstm networks,'' \emph{Proc. National Conference on Artificial Intelligence
  (AAAI)}, pp. 3697--3703, 2016.

\bibitem{zhang2017geometric}
S.~Zhang, X.~Liu, and J.~Xiao, ``On geometric features for skeleton-based
  action recognition using multilayer lstm networks,'' in \emph{Proc. IEEE
  Winter Conference on Applications of Computer Vision (WACV)}, 2017, pp.
  148--157.

\bibitem{liu2017sphereface}
W.~Liu, Y.~Wen, Z.~Yu, M.~Li, B.~Raj, and L.~Song, ``Sphereface: Deep
  hypersphere embedding for face recognition,'' in \emph{Proc. IEEE Conference
  on Computer Vision and Pattern Recognition (CVPR)}, vol.~1, 2017, p.~1.

\bibitem{simonyan2014two}
K.~Simonyan and A.~Zisserman, ``Two-stream convolutional networks for action
  recognition in videos,'' in \emph{Proc. Advances in Neural Information
  Processing Systems (NIPS)}, 2014, pp. 568--576.

\bibitem{lecun2015deep}
Y.~LeCun, Y.~Bengio, and G.~Hinton, ``Deep learning,'' \emph{Nature}, vol. 521,
  no. 7553, p. 436, 2015.

\bibitem{brox2011large}
T.~Brox and J.~Malik, ``Large displacement optical flow: descriptor matching in
  variational motion estimation,'' \emph{IEEE Trans. on Pattern Analysis and
  Machine Intelligence}, vol.~33, no.~3, pp. 500--513, 2011.

\bibitem{ojala2002multiresolution}
T.~Ojala, M.~Pietikainen, and T.~Maenpaa, ``Multiresolution gray-scale and
  rotation invariant texture classification with local binary patterns,''
  \emph{IEEE Trans. on Pattern Analysis and Machine Intelligence}, vol.~24,
  no.~7, pp. 971--987, 2002.

\bibitem{lee1988thirteen}
J.~Lee~Rodgers and W.~A. Nicewander, ``Thirteen ways to look at the correlation
  coefficient,'' \emph{The American Statistician}, vol.~42, no.~1, pp. 59--66,
  1988.

\bibitem{abadi2016tensorflow}
M.~Abadi, P.~Barham, J.~Chen, Z.~Chen, A.~Davis, J.~Dean, M.~Devin,
  S.~Ghemawat, G.~Irving, M.~Isard \emph{et~al.}, ``Tensorflow: a system for
  large-scale machine learning.'' in \emph{OSDI}, vol.~16, 2016, pp. 265--283.

\bibitem{kingma2014adam}
D.~P. Kingma and J.~Ba, ``Adam: A method for stochastic optimization,''
  \emph{arXiv preprint arXiv:1412.6980}, 2014.

\bibitem{Tian2005Real}
Z.~Tian and H.~Qin, ``Real-time driver's eye state detection,'' in \emph{Proc.
  IEEE International Conference on Vehicular Electronics and Safety (ICVES)},
  2005, pp. 285--289.

\bibitem{shou2016temporal}
Z.~Shou, D.~Wang, and S.-F. Chang, ``Temporal action localization in untrimmed
  videos via multi-stage cnns,'' in \emph{Proc. IEEE Conference on Computer
  Vision and Pattern Recognition (CVPR)}, 2016, pp. 1049--1058.

\bibitem{THUMOS14}
J.~Y.-G, L.~J, Z.~A.~Roshan, L.~I, P.~M, S.~M, and S.~R, ``Thumos challenge:
  Action recognition with a large number of classes.'' \url{
  http://crcv.ucf.edu/ICCV13-Action-Workshop/}, 2013.

\end{thebibliography}

\begin{IEEEbiography}[{\includegraphics[width=1in,height=1.25in,clip,keepaspectratio]{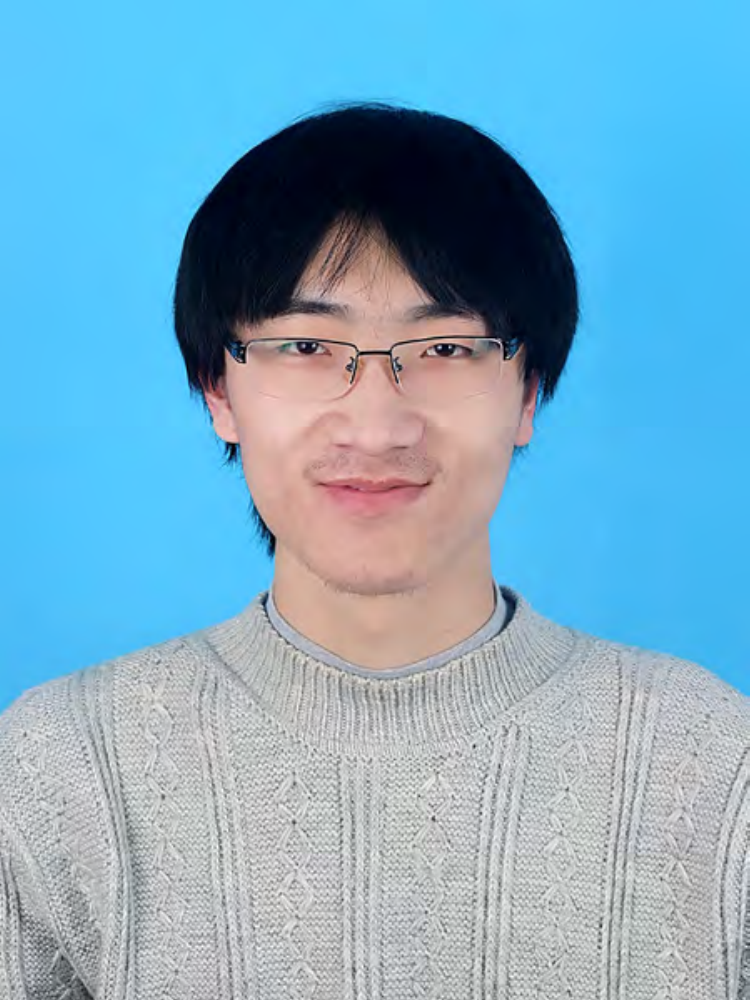}}]{Guilei Hu} received the B.S. degree from the Huazhong University of Science and Technology, China, in 2017. He is currently pursing the M.S. degree in the School of Artificial Intelligence and Automation at Huazhong University of Science and Technology. His research interests include human face parsing and eyeblink detection.
\end{IEEEbiography}	

\begin{IEEEbiography}[{\includegraphics[width=1in,height=1.25in,clip,keepaspectratio]{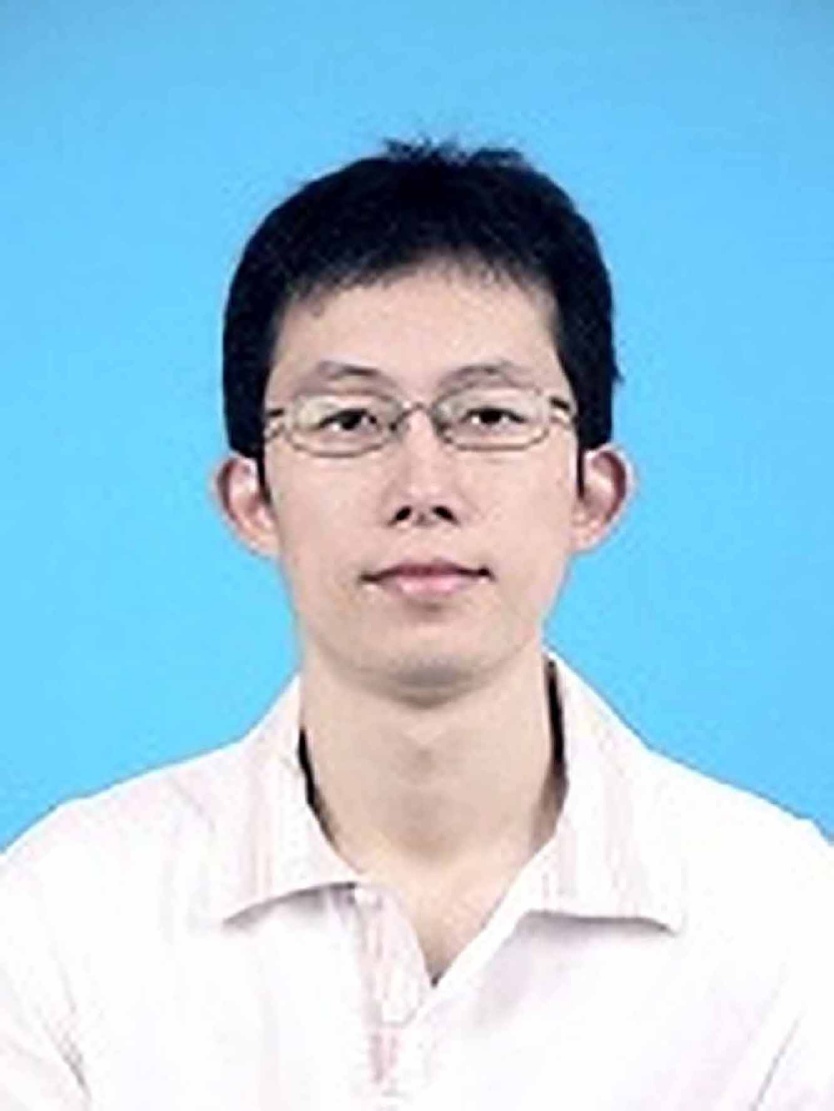}}]{Yang Xiao} received his BS, MS and PhD degrees from Huazhong University of Science and Technology, China. He is currently an associate professor in the School of Artificial Intelligence and Automation at Huazhong University of Science and Technology, China. Previously, he was ever the research fellow in the School of Computer Engineering and Institute of Media Innovation at Nanyang Technological University, Singapore. His research interests involve computer vision, image processing and machine learning.
\end{IEEEbiography}	

\begin{IEEEbiography}[{\includegraphics[width=1in,height=1.25in,clip,keepaspectratio]{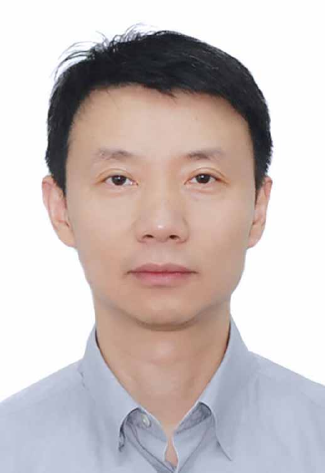}}]{Zhiguo Cao} is a professor of School of Artificial Intelligence and Automation in Huazhong University of Science and Technology. He received his BS and MS degrees in communication and information System from the University of Electronic Science and Technology of China, and his PhD degree in Pattern Recognition and Intelligent System from Huazhong University of Science and Technology. His research interests spread across image understanding and analysis, depth information extraction, 3d video processing, motion detection and human action analysis. His research results, which have published dozens of papers at international journals and prominent conferences, have been applied to automatic observation system for crop growth in agricultural, for weather phenomenon in meteorology and for object recognition in video surveillance system based on computer vision.
\end{IEEEbiography}

\begin{IEEEbiography}[{\includegraphics[width=1in,height=1.25in,clip,keepaspectratio]{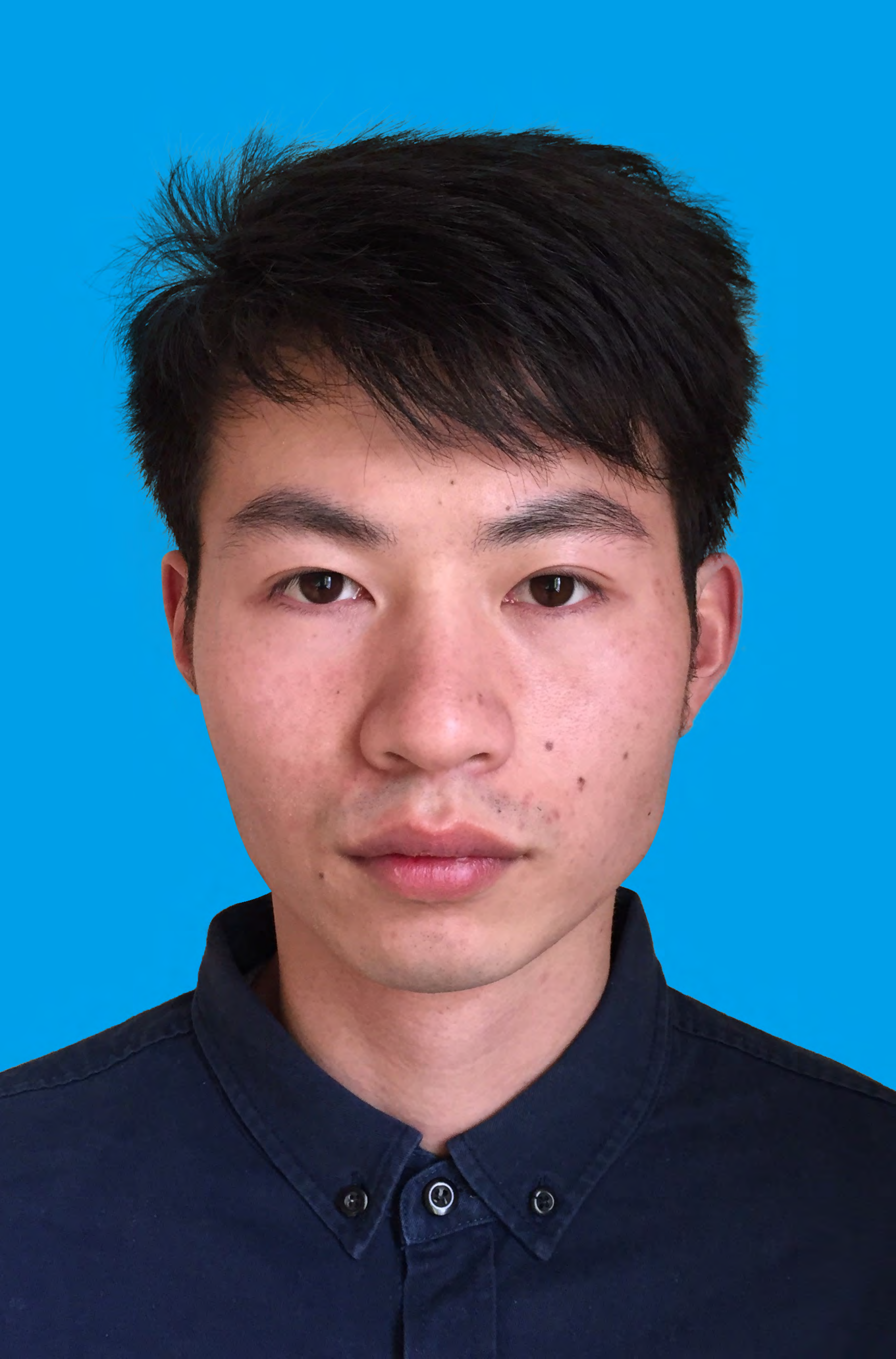}}]{Lubin Meng} is currently pursing the B.S. degree at Huazhong University of Science and Technology, Wuhan, China. His research interests include adversarial examples and domain adaptation.
\end{IEEEbiography}	

\begin{IEEEbiography}[{\includegraphics[width=1in,height=1.25in,clip,keepaspectratio]{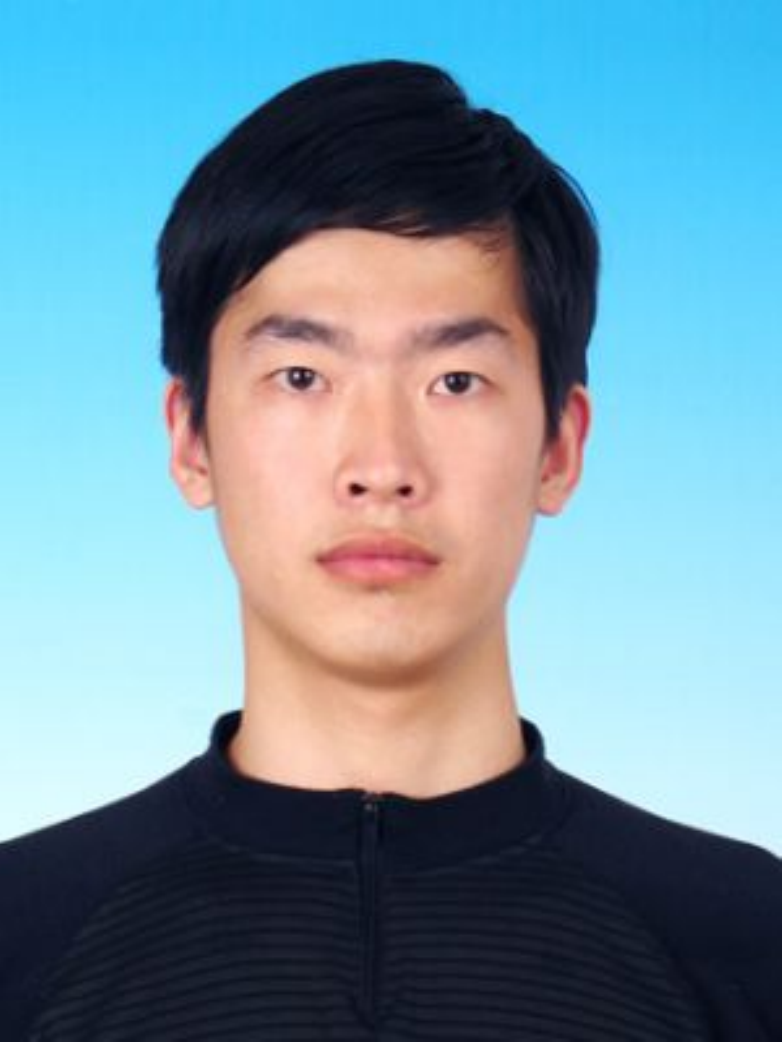}}]{Zhiwen Fang} received his B.S. and M.S. degrees in the Automation School of Beihang University, and his PhD degree from Huazhong University of Science and Technology. His research interests include object detection, object tracking, anomaly detection and machine learning.
\end{IEEEbiography}	

\begin{IEEEbiography}[{\includegraphics[width=1in,height=1.25in,clip,keepaspectratio]{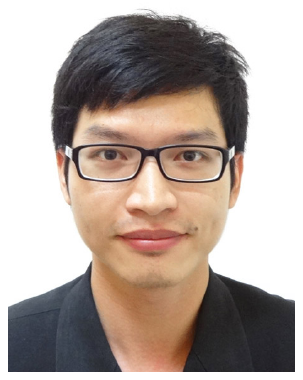}}]{Joey Tianyi Zhou} received the Ph.D. degree in computer science from Nanyang Technological University, Singapore, in 2015.
He is currently a Scientist with the Institute of High Performance Computing, Research Agency for Science, Technology, and Research, Singapore.

Dr. Zhou was a recipient of the Best Poster Honorable Mention at ACML 2012, the Best Paper Award from the BeyondLabeler Workshop on IJCAI 2016, the Best Paper Nomination at ECCV 2016, and the NIPS 2017 Best Reviewer Award. He has served as an Associate Editor for IEEE Access, a Guest Editor for IET Image Processing.
\end{IEEEbiography}

\begin{IEEEbiography}[{\includegraphics[width=1in,height=1.25in,clip,keepaspectratio]{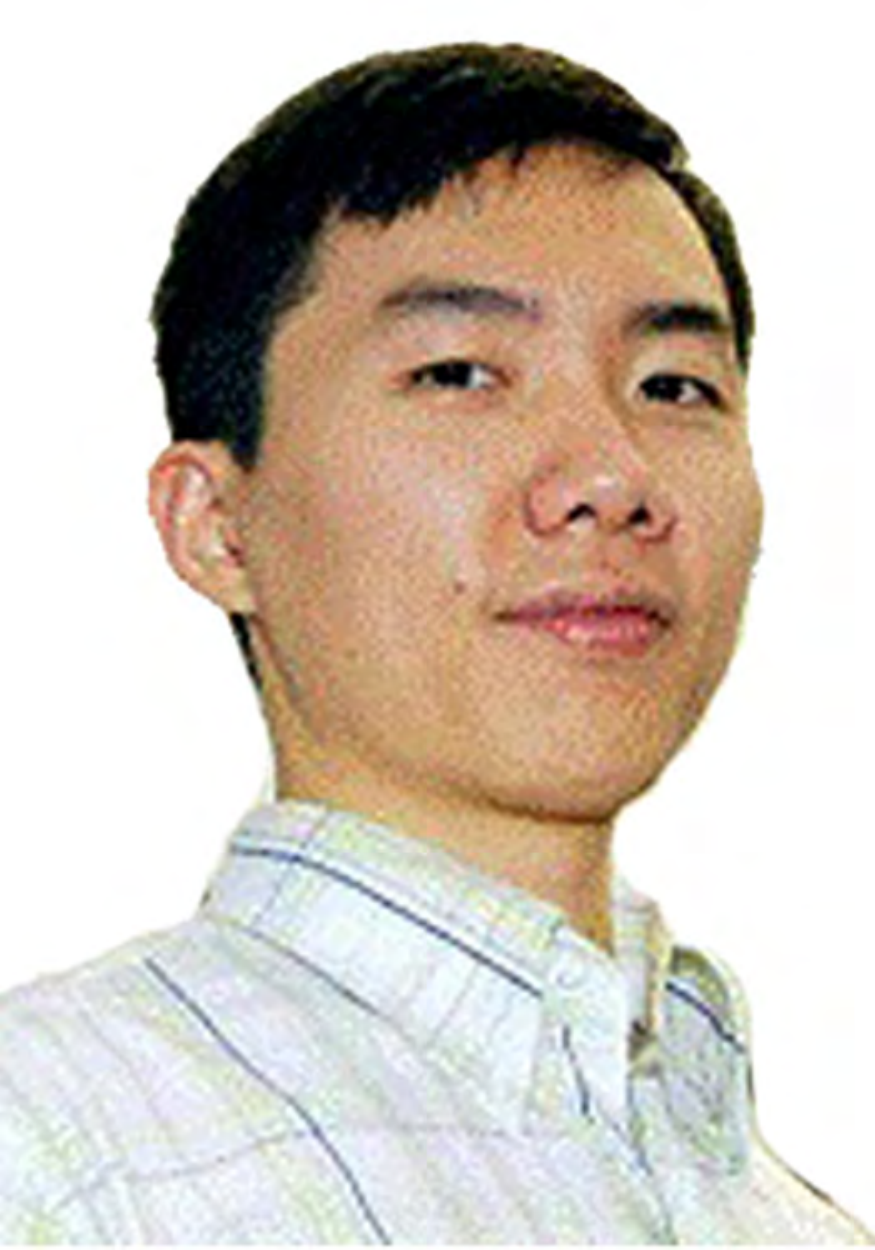}}]{Junsong Yuan} (M08-SM14) received Ph.D. from Northwestern University in 2009 and M.Eng. from National University of Singapore in 2005. Before that, he graduated from the Special Class for the Gifted Young of Huazhong University of Science and Technology, P.R.China, in 2002. He is currently an associate professor and director of visual computing lab at Computer Science and Engineering Department of University at Buffalo, the State University of New York, USA. Before that he was an associate professor at School of Electrical and Electronics Engineering (EEE), Nanyang Technological University (NTU), Singapore. He received Best Paper Award from IEEE Trans. on Multimedia, Doctoral Spotlight Award from IEEE Conf. on Computer Vision and Pattern Recognition (CVPR09), Nanyang Assistant Professorship from NTU, and Outstanding EECS Ph.D. Thesis award from Northwestern University. He is currently Senior Area Editor of Journal of Visual Communication and Image Representation (JVCI), Associate Editor of IEEE Trans. on Image Processing (T-IP), IEEE Trans. on Circuits and Systems for Video Technology (T-CSVT), and served as Guest Editor of International Journal of Computer Vision (IJCV). He was Program Co-chair of ICME'18 and VCIP'15, and Area Chair of CVPR, ACM MM, ACCV, WACV, ICPR, ICIP etc. He is a fellow of the International Association of Pattern Recognition (IAPR).
\end{IEEEbiography}		

\end{document}